\documentclass[mlmain]{jmlr}

\usepackage{multirow}
\usepackage{booktabs}
\usepackage{longtable}
\usepackage{threeparttable}
\usepackage{wrapfig}

\jmlrvolume{}
\jmlrpages{}
\firstpageno{1}
\jmlryear{2026}
\jmlrworkshop{}
\jmlrproceedings{Preprint}{Preprint}
\makeatletter
\gdef\@editor{}
\makeatother

\title[SAMBA]{SAMBA: Toward an EEG Foundation Model via Differential Mamba and Coordinate-Driven Spatial Embedding}

\author{
\Name{Jiazhen Hong} \Email{jiazhen@emotiv.com}\\
\addr Emotiv Research, California, USA
\AND
\Name{Geoffrey Mackellar} \Email{geoff@emotiv.com}\\
\addr Emotiv Research, Sydney, Australia
\AND
\Name{Soheila Ghane}\thanks{Work performed while the author was affiliated with Emotiv.}
\Email{soheila@google.com}\\
\addr Google, Melbourne, Australia
}

\begin{document}
\maketitle

%---------JH----------
% TL;DR
% SAMBA is a self-supervised framework toward long-sequence EEG foundation modeling, integrating coordinate-driven spatial embedding and differential Mamba for scalable and efficient EEG representation learning.

%%
%% The "author" command and its associated commands are used to define
%% the authors and their affiliations.
%% Of note is the shared affiliation of the first two authors, and the
%% "authornote" and "authornotemark" commands
%% used to denote shared contribution to the research.

% ------------------------------------------------------------------
% \author{Jiazhen Hong}
% \affiliation{%
%   \institution{Emotiv Research}
%   \city{Melbourne}
%   \country{Australia}
% }
% \email{jiazhen@emotiv.com}
% % ------------------------------------------------------------------
% \author{Geoffrey Mackellar}
% \affiliation{%
%   \institution{Emotiv Research}
%   \city{Sydney}
%   \country{Australia}
% }
% \email{geoff@emotiv.com}
% % ------------------------------------------------------------------
% \author{Soheila Ghane}
% \affiliation{%
%   \institution{Emotiv Research}
%   \city{Melbourne}
%   \country{Australia}
% }
% \email{soheila@emotiv.com}
% ------------------------------------------------------------------

%%
%% The abstract is a short summary of the work to be presented in the
%% article.
\begin{abstract}
Modeling long-context electroencephalography (EEG) signals is critical for learning generalizable EEG representations, as EEG recordings are acquired at high sampling rates and often span extended durations that capture evolving neural states beyond short task-specific windows.
Transformer-based models have shown promise in modeling short sequences of a few seconds; however, their quadratic complexity limits scalability to longer contexts. 
Moreover, substantial variability in electrode montages across datasets and recording devices poses a major challenge for building EEG models that generalize across domains.
We propose \textit{SAMBA}, a self-supervised learning framework with a Mamba-based U-shaped encoder-decoder architecture, which effectively captures long-range temporal dependencies and montage variability in EEG data. Leveraging the inherent ability of Mamba in processing long context sizes, we introduce: (1) \textit{Temporal Semantic Random Masking} for semantic-level sequence reconstruction, (2) a \textit{Multi-Head Differential Mamba} module to suppress redundancy and emphasize salient temporal structures, and (3) a \textit{Coordinate-Driven Spatial Embedding} that learns unified embeddings in a three-dimensional Euclidean space, enabling robustness across devices.
Experiments on thirteen EEG datasets across diverse tasks, electrode configurations, and sequence durations demonstrate that SAMBA achieves competitive or improved performance compared to recent state-of-the-art methods, while maintaining low memory consumption and efficient inference. The learned spatial weight maps exhibit close alignment with task-relevant neurophysiological regions, highlighting the interpretability of SAMBA. 
The code is available at: \url{https://github.com/Jiazhen-Hong/SAMBA}
\end{abstract}

%%
%% The code below is generated by the tool at http://dl.acm.org/ccs.cfm.
%% Please copy and paste the code instead of the example below.
%%
% \begin{CCSXML}
% <ccs2012>
%  <concept>
%   <concept_id>00000000.0000000.0000000</concept_id>
%   <concept_desc>Do Not Use This Code, Generate the Correct Terms for Your Paper</concept_desc>
%   <concept_significance>500</concept_significance>
%  </concept>
%  <concept>
%   <concept_id>00000000.00000000.00000000</concept_id>
%   <concept_desc>Do Not Use This Code, Generate the Correct Terms for Your Paper</concept_desc>
%   <concept_significance>300</concept_significance>
%  </concept>
%  <concept>
%   <concept_id>00000000.00000000.00000000</concept_id>
%   <concept_desc>Do Not Use This Code, Generate the Correct Terms for Your Paper</concept_desc>
%   <concept_significance>100</concept_significance>
%  </concept>
%  <concept>
%   <concept_id>00000000.00000000.00000000</concept_id>
%   <concept_desc>Do Not Use This Code, Generate the Correct Terms for Your Paper</concept_desc>
%   <concept_significance>100</concept_significance>
%  </concept>
% </ccs2012>
% \end{CCSXML}

% \begin{keywords}
% Electroencephalography (EEG), Long-Sequence Modeling, EEG Self-Supervised Learning, Differential Mamba, EEG Spatial Embedding
% \end{keywords}
\begin{keywords}
Electroencephalography, Long-Sequence Modeling, Self-Supervised Learning, Differential Mamba, Spatial Embedding
\end{keywords}
%% A "teaser" image appears between the author and affiliation
%% information and the body of the document, and typically spans the
%% page.
% \begin{teaserfigure}
%   \includegraphics[width=\textwidth]{sampleteaser}
%   \caption{Seattle Mariners at Spring Training, 2010.}
%   \Description{Enjoying the baseball game from the third-base
%   seats. Ichiro Suzuki preparing to bat.}
%   \label{fig:teaser}
% \end{teaserfigure}

% \received{20 February 2007}
% \received[revised]{12 March 2009}
% \received[accepted]{5 June 2009}

\section{Introduction}
Long-sequence electroencephalogram (EEG) modeling is increasingly required in real-world applications that involve either high sampling rates or extended recording durations. For example, EEG signals may be acquired at very high temporal resolutions in steady-state visual evoked potential (SSVEP) paradigms~\cite{li2023design}, or span tens of seconds to hours in applications such as sleep analysis~\cite{campbell2009eeg}, emotion recognition~\cite{zheng2018emotionmeter}, and long-term cognitive or fatigue monitoring~\cite{zheng2017multimodal}. In both cases, learning effective representations from long EEG sequences remains challenging.

\textit{(1) High memory usage in long-sequence modeling.}
Transformer-based SSL approaches such as MAEEG~\cite{chien2211maeeg}, BENDR~\cite{kostas2021bendr}, BIOT~\cite{yang2024biot}, EEG2Rep~\cite{eeg2rep2024}, LabraM~\cite{jiang2024large}, and EEGPT~\cite{wangeegpt} have shown strong performance on short EEG segments. However, in long-term EEG monitoring tasks, even modest downsampling to 128 Hz yields extremely long input sequences (e.g., 12,800 time steps for a 100-second recording). Despite architectural variants aimed at capturing long-range dependencies, Transformers inherently struggle with such sequences due to their \(\mathcal{O}(n^2)\) complexity~\cite{gu2021efficiently}. This not only limits the feasibility of long-sequence modeling, but also leads to excessive memory usage and slow inference. This motivates an efficient framework based on structured state space models (SSMs)~\cite{gu2021efficiently}.

% \begin{figure}
%   \centering
%   \includegraphics[width=0.7\linewidth]{Figure/Fig1.pdf}
%   \vspace{-0.15in}
%   \caption{Channel-handling assumptions across heterogeneous EEG electrode layouts.}\label{fig:montages}
% \end{figure}

\begin{wrapfigure}[16]{r}{0.7\textwidth}
  \centering
  \setlength{\abovecaptionskip}{2pt}
  \includegraphics[width=\linewidth]{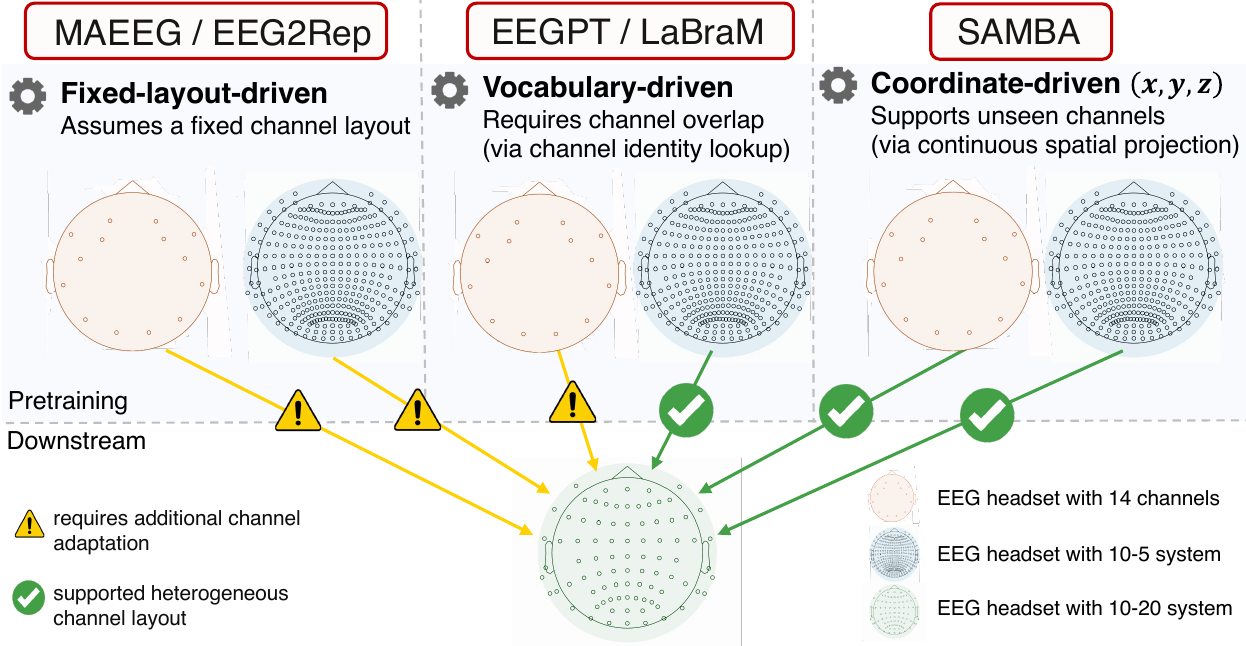}
  \caption{Channel handling across EEG montages.}
  \label{fig:montages}
\end{wrapfigure}

\textit{(2) Diverse EEG headsets and montages.}  
EEG signals are typically represented as a matrix \( \mathbf{X} \in \mathbb{R}^{C \times T} \), where \(C\) denotes the number of channels (electrodes) and \(T\) the number of time points. In practice, both vary widely across datasets due to differences in equipment, montage configurations, and application-specific requirements. For example, P300 speller brain-computer interfaces (BCIs) may use only 8 or 16 channels~\cite{hong2026chatbciassist}, whereas EEG source localization analysis requires high-density 128-channel setups~\cite{haddad2018source}. Even when channel names are consistent, their spatial coordinates \((x, y, z)\) may differ between devices or cap systems. 
This heterogeneity limits model generalization and cross-domain transferability (Fig.~\ref{fig:montages}).
For instance, EEG2Rep~\cite{eeg2rep2024} and MAEEG~\cite{chien2211maeeg} assume consistent channel layouts between pretraining and downstream tasks, and require additional channel adaptation when electrode configurations differ. 
EEGPT~\cite{wangeegpt} and LaBraM~\cite{jiang2024large} adopt vocabulary-driven spatial embeddings based on predefined electrode sets (e.g., 58 channels for EEGPT and the 10--20 system for LaBraM). As a result, they typically require additional channel mapping to handle unseen electrode configurations.

\textit{(3) Empirical fixed-length EEG segmentation.}   
Existing methods often constrain temporal modeling by applying empirical fixed-length window or patch segmentation. For instance, EEGPT~\cite{wangeegpt} and LaBraM~\cite{jiang2024large} adopt local spatio-temporal patching strategies with varied channel configurations, but both use a fixed one-second non-overlapping time window as the patch unit. These designs may overlook temporal semantics between patches and are suboptimal with data segments shorter than one second. In time-locked event-related potential (ERP) paradigms, for example, the P300 component ($\sim 300 ms$) reflects attention-related target detection~\cite{hong2024p3t}, while N170 ($\sim 170 ms$) marks early face perception over occipito-temporal regions~\cite{caharel2021n170}. Such components are highly time-sensitive and often span sub-second durations, making them vulnerable to distortion or loss under fixed-length segmentation. 

\textit{(4) Learning robust representations under low SNR and subject dependence.}  
EEG signals are characterized by a low signal-to-noise ratio (SNR) and non-stationary properties~\cite{hong2022deep}, making robust representation learning particularly challenging.  
Despite the emergence of various deep learning paradigms for EEG representation learning, many researchers still prefer to design handcrafted features for subject-specific models~\cite{wang2023st, ma2021capsule}. 
In contrast, BIOT~\cite{yang2024biot} adopts a unified model across subjects but relies on spectral-domain transformations (e.g., the fast Fourier transform), which may compromise temporal information. 

%Contribution
To address these challenges, we introduce \textbf{SAMBA}, a U-shaped SSL framework based on Mamba, which integrates: \textit{Temporal Semantic Random (TSR) Masking}, \textit{Coordinate-Driven Spatial Embedding (CDSE)}, and \textit{Multi-Head Differential Mamba (MDM)} modules. The main contributions are summarized as follows.

\noindent\textbullet \textbf{Memory-Efficient Long-Sequence Modeling:}  
SAMBA leverages Mamba2~\cite{dao2024transformers} structured state space models with linear-time complexity, avoiding the quadratic cost of attention mechanisms. This enables the model to handle long EEG recordings in tasks such as abnormality detection without excessive memory usage.

% \item \textbf{Montage-Agnostic Spatial Embedding:} 
\noindent\textbullet \textbf{Spatial Compatibility with Diverse EEG Montages:}
SAMBA employs CDSE to leverage electrode \((x, y, z)\) coordinates, enabling generalization across diverse EEG montages without relying on electrode names. For example, a model pretrained on the 10--20 system can be applied to unseen configurations, such as 14-channel montages or the 10--5 system with over 300 electrodes (Fig.~\ref{fig:montages}).

\noindent\textbullet \textbf{Temporal Compatibility with Various EEG durations:}  
SAMBA integrates Mamba2 layers into a hierarchical encoder-decoder to efficiently capture local and global temporal dependencies.
Its components are designed to be position-agnostic and invariant to input sequence length, enabling SAMBA to handle EEG sequences ranging from 256 to 12,800 time steps without architectural modifications while maintaining robust performance.

\noindent\textbullet \textbf{Effective EEG Representation Learning:}  
SAMBA employs TSR masking to capture semantic structures across diverse time segments and MDM to suppress redundancy and enhance salient temporal features by contrasting dual Mamba2 pathways per head. A combined temporal and spectral loss further preserves information in both domains.
Representation visualizations across montage and duration variations are provided in Appendix~\ref{app:moved-tsne}.

\section{Method} \label{sec:method}
\begin{figure}
  \centering
  \includegraphics[width=\linewidth]{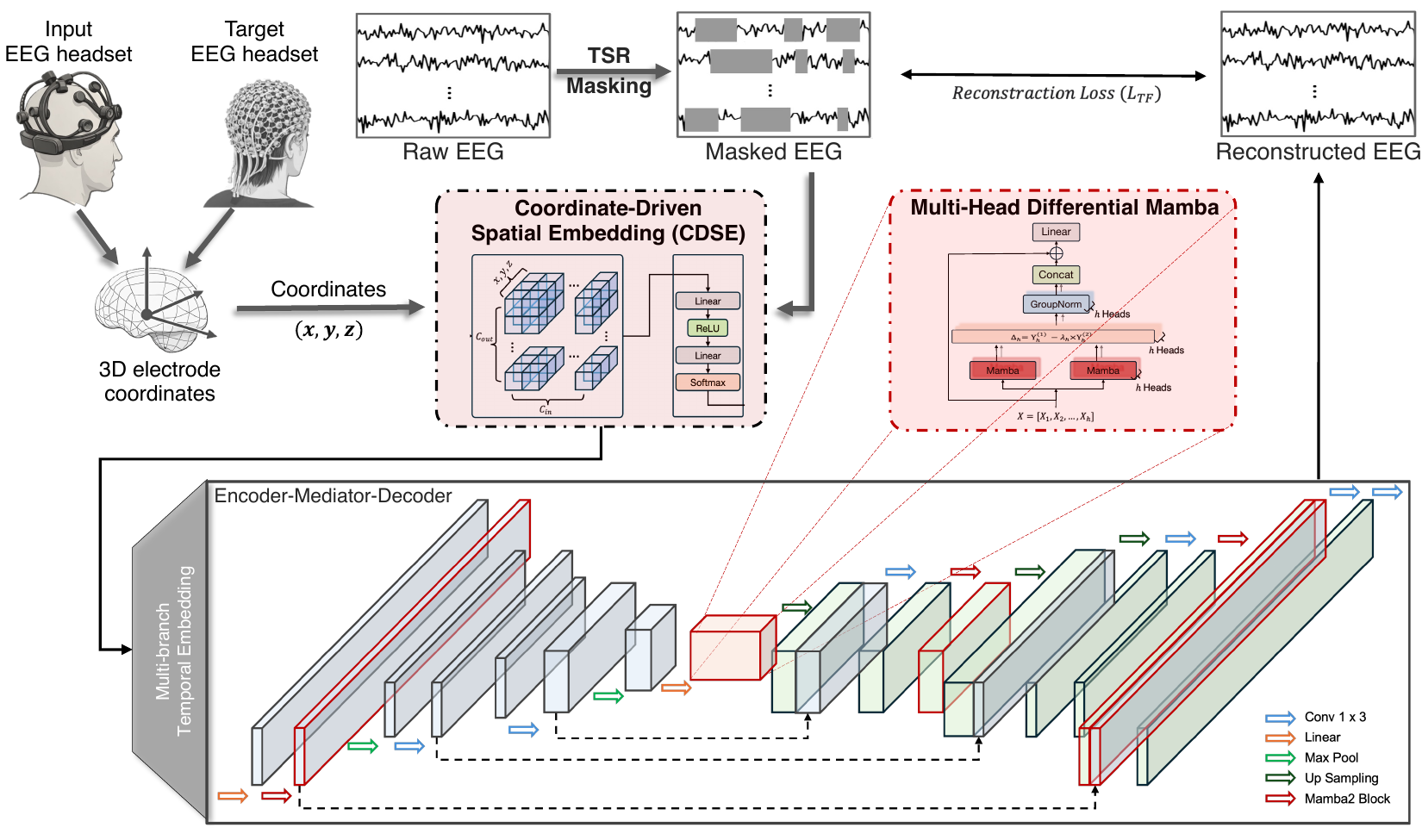}
  \vspace{-0.15in}
  \caption{SAMBA Architecture.}\label{fig:1}
\end{figure}

\paragraph{Architecture Overview:}
Figure~\ref{fig:1} illustrates the overall architecture of SAMBA. Given an input EEG sequence \( X \in \mathbb{R}^{B \times C_{\text{in}} \times T} \), where \( B \) is the batch size, \( C_{\text{in}} \) the number of input channels, and \( T \) the number of time steps, the sequence is first processed by the TSR masking strategy (Appendix~\ref{sec:tsr}) with a 50\% time mask and 0\% channel mask. The masked input is then projected into the standard montage (using the 10--20 system in this work) based on \((x, y, z)\) coordinates of the electrodes using the CDSE (Section~\ref{sec:cdse}), followed by a multi-branch temporal embedding with parallel 1D convolutions of kernel sizes (e.g., 1, 3, and 7) to enrich local temporal features at multiple resolutions.

The encoder consists of three stages of increasing depth and decreasing temporal resolution. The first stage applies a linear projection followed by a Mamba2 block~\cite{dao2024transformers} to capture fine-grained temporal dynamics. The second and third stages utilize 1D convolutions and max pooling to downsample the temporal dimension while expanding the feature dimension, enabling subsequent Mamba blocks to model longer-range dependencies over compact feature representations. This hierarchical design supports sequences of varying length without requiring fixed-size temporal windows or architectural modifications.

At the bottleneck, a Multi-head Differential Mamba (Section~\ref{sec:mdm}) module is used as a mediator to contrast parallel state-space dynamics across multiple heads and suppress noise.

The decoder mirrors the encoder with symmetric structure and performs upsampling via parameter-free linear interpolation instead of transposed convolutions. This avoids checkerboard artifacts~\cite{sugawara2019checkerboard} and better preserves the continuity of EEG signals. Each upsampled feature map is refined by a Mamba2 block to restore long-range temporal patterns. 
To preserve both time-domain waveform and frequency-domain structure, we adopt a Time-Frequency loss function (\( \mathcal{L}_{\text{TF}} \)), which combines an \( L_1 \) loss in the time domain with a spectral alignment loss based on the discrete Fourier transform. Further details are provided in Appendix~\ref{app:loss}. This joint objective encourages SAMBA to maintain fidelity in both temporal shape and frequency content, supporting robust EEG modeling.

\subsection{Coordinate-Driven Spatial Embedding (CDSE)}\label{sec:cdse}

\begin{figure}
  \centering
  \includegraphics[width= 0.8\linewidth]{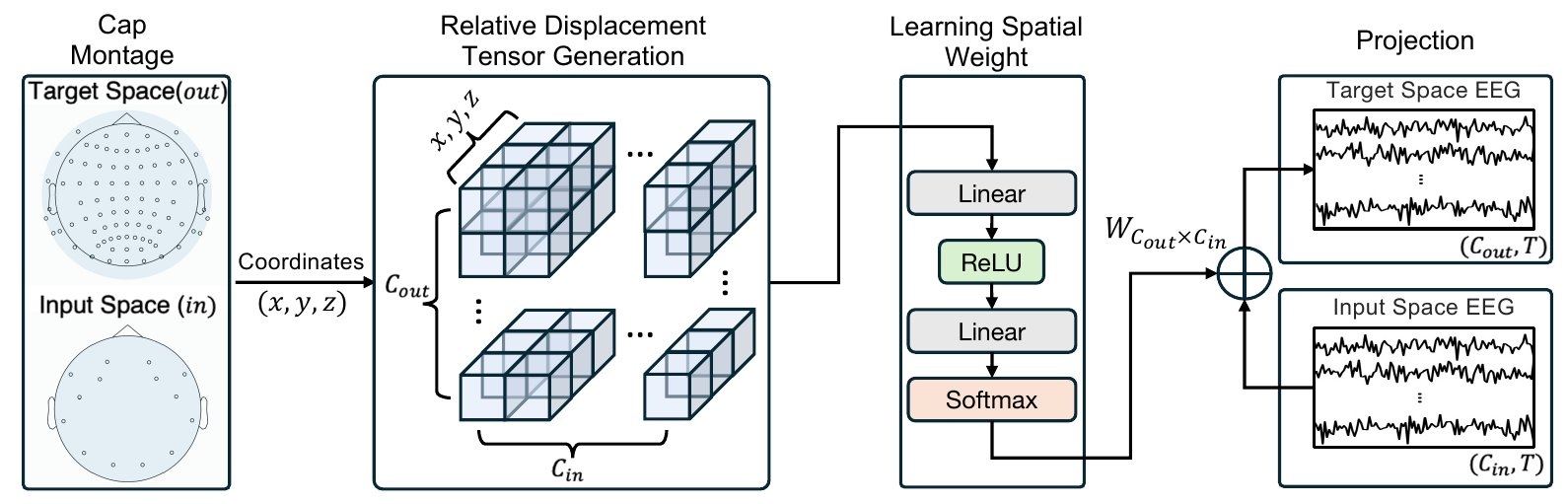}
  \vspace{-0.15in}
  \caption{CDSE projects EEG using spatial weights derived from relative 3D coordinates.}\label{fig:3}
\end{figure}

% Electrode layouts vary across EEG datasets due to differences in headsets, montages, and clinical objectives. 
% Electrode layouts vary across EEG datasets due to differences in devices and requirements.
Electrode layouts vary across EEG datasets and devices. To address the variability, we introduce Coordinate-Driven Spatial Embedding (CDSE), which uses 3D electrode coordinates to map input signals into a standard montage. 
% Figure~\ref{fig:3} illustrates CDSE using the standard 10--20 system montage~\cite{gramfort2013meg} as the target space and Emotiv's 14-channel montage~\cite{williams2023crowdsourced} as an input space. 
Figure~\ref{fig:3} illustrates CDSE with the standard 10--20 system~\cite{gramfort2013meg} as the target montage and Emotiv's 14-channel layout~\cite{williams2023crowdsourced} as the input.
Given input EEG \( \mathbf{X} \in \mathbb{R}^{B \times C_{\text{in}} \times T} \), let \( \mathbf{P}_{\text{in}} \in \mathbb{R}^{C_{\text{in}} \times 3} \) and \( \mathbf{P}_{\text{out}} \in \mathbb{R}^{C_{\text{out}} \times 3} \) denote the 3D coordinates of input and target electrodes. The relative displacement between each pair is:
$\Delta \mathbf{P}_{ij} = \mathbf{P}_{\text{out}, i} - \mathbf{P}_{\text{in}, j}$.
The unnormalized spatial weights are generated by a multilayer perceptron (MLP):
\begin{equation}
w_{ij} = \phi_\theta(\Delta \mathbf{P}_{ij}) = \mathbf{W}_2 \cdot \mathrm{ReLU}(\mathbf{W}_1 \Delta \mathbf{P}_{ij} + \mathbf{b}_1) + \mathbf{b}_2,
\end{equation}
followed by softmax normalization across input channels:
\begin{equation}
\tilde{w}_{ij} = \frac{\exp(w_{ij})}{\sum_{k=1}^{C_{\text{in}}} \exp(w_{ik})}.
\end{equation}
The projected signal at target channel \( i \) is computed as a weighted sum over input channels:
\begin{equation}
\mathbf{X}'_{b,i,t} = \sum_{j=1}^{C_{\text{in}}} \tilde{w}_{ij} \cdot \mathbf{X}_{b,j,t},
\end{equation}

\begin{wrapfigure}{r}{0.48\linewidth}
  \centering
  \vspace{-10pt}
  \includegraphics[width=\linewidth]{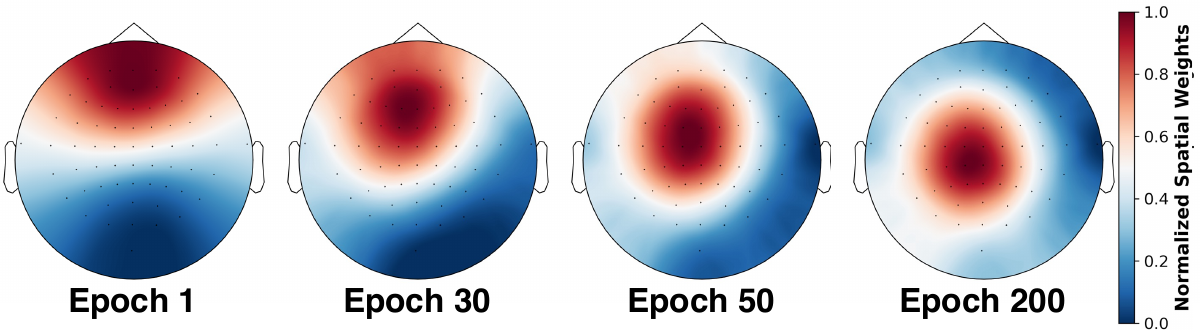}
  \vspace{-10pt}
  \caption{Evolution of spatial weights.}
  \label{fig:mi_topo}
  \vspace{-10pt}
\end{wrapfigure}
where \( b \) indexes the batch, \( t \) the time step, \( j \) the input channel, and \( i \) the target channel. Figure~\ref{fig:mi_topo} illustrates the topographic evolution of learned spatial weights during pretraining on the PhysionetMI dataset (motor imagery task). Darker red regions indicate higher weights, consistently localized around the motor cortex~\cite{hong2022deep}.
% where \(b\) and \(t\) index the batch and time step, while \(j\) and \(i\) denote the input and target channels, respectively. Figure~\ref{fig:mi_topo} shows the evolution of learned spatial weights during pretraining on motor imagery and eyes open/closed tasks. Darker red indicates higher weights, with task-dependent spatial patterns emerging over training epochs.
% \begin{figure}
%   \centering
%   \includegraphics[width=0.5\linewidth]{Figure/Fig6-MITopo.pdf}
%   % \includegraphics[width=\linewidth]{Figure/Fig6.pdf}
%   \vspace{-0.15in}
%   \caption{Topographic evolution of spatial weights}
%   \label{fig:mi_topo}
% \end{figure}

\subsection{Multi-head Differential Mamba (MDM)}\label{sec:mdm}
\begin{figure*}
  \centering
  \includegraphics[width=0.8\linewidth]{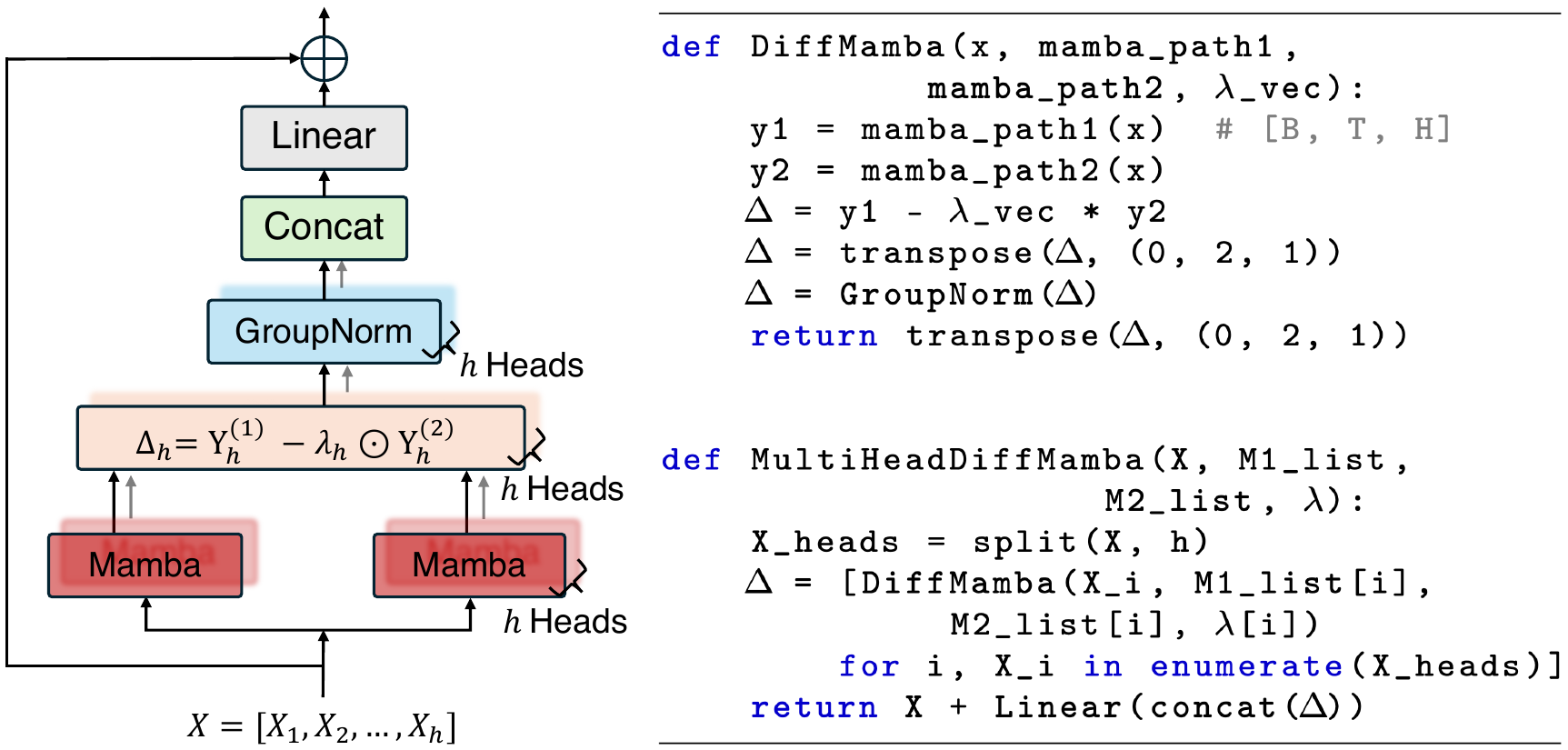}
  \vspace{-0.15in}
  \caption{Multi-head Differential Mamba (MDM) block.}\label{fig:4}
\end{figure*}
Building on the motivation for adopting Mamba2 (Appendix~\ref{app:mamba}), we extend its capabilities by proposing the Multi-head Differential Mamba (MDM) module to enhance EEG representations at the encoder-decoder bottleneck. Inspired by differential attention~\cite{ye2024differential}, which suppresses noise by contrasting dual attention maps, MDM contrasts two Mamba2 outputs within each head, enabling continuous temporal processing without softmax-based attention.

Unlike differential attention operating in similarity space ($QK^\top$), each head in MDM processes the input with two independently parameterized Mamba2 modules, enabling diverse dynamic representations. As shown in Figure~\ref{fig:4}, given input \( X \in \mathbb{R}^{B \times T \times D} \), where \( B \), \( T \), and \( D \) are batch size, sequence length, and feature dimension respectively, we divide \( X \) along the feature dimension into \( H \) heads:
\begin{equation}
X = [X_1, \dots, X_H], \quad X_h \in \mathbb{R}^{B \times T \times d}, \quad d = D / H.
\end{equation}

Each head applies two separate Mamba2 modules:
\begin{equation}
Y_h^{(1)} = \mathrm{Mamba}_h^{(1)}(X_h), \quad Y_h^{(2)} = \mathrm{Mamba}_h^{(2)}(X_h).
\end{equation}

A learnable scaling vector \( \lambda_h \in \mathbb{R}^{d} \) modulates the difference between the two outputs, followed by per-head GroupNorm:
\begin{equation}
\widetilde{\Delta}_h = \mathrm{GroupNorm}_h(Y_h^{(1)} - \lambda_h \odot Y_h^{(2)}).
\end{equation}

The outputs from all heads are concatenated, linearly projected, and added to the original input via a residual connection:
\begin{equation}
Y = X + \mathrm{Linear}(\mathrm{Concat}(\widetilde{\Delta}_1, \dots, \widetilde{\Delta}_H)).
\end{equation}

% \paragraph{Residual Connection Justification.} 
% While the original Differential Transformer~\cite{ye2024differential} excludes residuals to isolate contrastive information, we retain the skip connection to preserve meaningful EEG components such as slow trends and rhythmic patterns. 

\section{Experimental Setup and Results} \label{exp}\label{results}
We evaluate SAMBA in terms of (1) \textit{learnability} through in-domain performance, (2) \textit{transferability} across datasets and domains, and (3) \textit{scalability} toward foundation models. Ablation studies are provided in Appendix~\ref{app:moved-ablation}, while additional details on the datasets, preprocessing, implementation, and evaluation metrics are provided in Appendix~\ref{app:moved-experimental}.

\begin{figure}
  \centering
  \includegraphics[width=\linewidth]{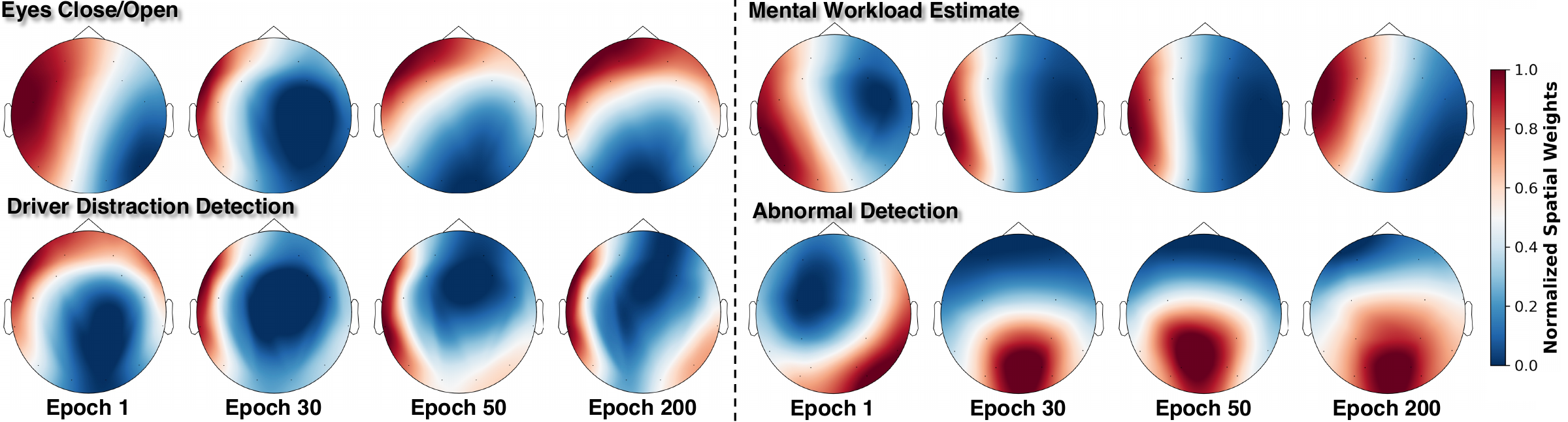}
  \vspace{-0.15in}
  \caption{Evolution of spatial weights during in-domain pretraining.}
  \label{fig:8}
\end{figure}

\subsection{In-Domain Learning} \label{re:indomain}
We evaluate in-domain learning from two perspectives: the spatial patterns captured by CDSE and the quality of the learned representations.

\paragraph{Spatial Learnability} Figure~\ref{fig:8} visualizes the spatial embedding weights learned across four EEG tasks. 
The weights evolve from diffuse patterns at epoch 1 to more stable, task-specific patterns by epochs 50--200. The learned emphasis differs across tasks, covering frontal regions for eye state, temporal regions for driver distraction, left-frontal regions for mental workload, and occipital regions for abnormal EEG detection. These patterns qualitatively align with prior neurophysiological findings~\cite{hoffmann2008correction,li2023drivers,berretz2022acute,lopez2015automated}, suggesting that CDSE captures task-relevant spatial structure.

\paragraph{Representation learnability.}
We evaluate downstream representations on the Emotiv datasets and TUAB, following the benchmarks used by EEG2Rep~\cite{eeg2rep2024}, EEGPT~\cite{wangeegpt}, and LaBraM~\cite{jiang2024large}.
On the Emotiv datasets, SAMBA achieves the best linear-probing results across all tasks and metrics (Table~\ref{tab:2}). After fine-tuning, it obtains the highest AUROC on all three tasks and competitive accuracy. Training from random initialization performs substantially worse, confirming the benefit of pretraining.

\begin{table}
\centering
\caption{Performance on the Emotiv datasets.}\label{tab:2}
\resizebox{0.7\linewidth}{!}{
\begin{tabular}{l cc cc cc}
\toprule
\multirow{2}{*}{\textbf{Models}} & \multicolumn{2}{c}{\textbf{Crowdsourced}} & \multicolumn{2}{c}{\textbf{DriverDistraction}} & \multicolumn{2}{c}{\textbf{STEW}}  \\ 
\cline{2-7} 
& \multicolumn{1}{c}{\textbf{ACC}} & \textbf{AUROC} & \multicolumn{1}{c}{\textbf{ACC}} & \textbf{AUROC} & \multicolumn{1}{c}{\textbf{ACC}} & \textbf{AUROC}  \\ 
\hline

BENDR & 70.46$\pm$4.14 & 0.7056$\pm$0.04 & 68.40$\pm$3.08 & 0.5521$\pm$0.03 & 63.03$\pm$1.07 & 0.6303$\pm$0.01 \\
MAEEG & 75.21$\pm$2.11 & 0.7501$\pm$0.02 & 68.37$\pm$2.60 & 0.5529$\pm$0.02 & 67.99$\pm$1.86 & 0.6858$\pm$0.02 \\
BIOT & 76.23$\pm$4.56 & 0.7633$\pm$0.04 & 63.93$\pm$1.28 & 0.6333$\pm$0.01 & 67.54$\pm$2.08 & 0.6769$\pm$0.04 \\
EEG2Rep & 81.66$\pm$2.93 & 0.8167$\pm$0.03 & 76.88$\pm$2.55 & 0.6559$\pm$0.02 & 69.04$\pm$1.04 & 0.6910$\pm$0.01 \\
\underline{SAMBA (Linear)} & \textbf{86.72$\pm$0.02} & \textbf{0.9336$\pm$0.00} & \textbf{77.97$\pm$0.90} & \textbf{0.7092$\pm$0.01} & \textbf{70.62$\pm$0.03} & \textbf{0.7704$\pm$0.00} \\ 
\addlinespace[1pt]
\hline
BENDR & 83.78$\pm$2.35 & 0.8380$\pm$0.03 & 74.31$\pm$2.38 & 0.5986$\pm$0.03 & 69.74$\pm$2.11 & 0.6977$\pm$0.02 \\
MAEEG & 86.75$\pm$3.50 & 0.8621$\pm$0.03 & 74.58$\pm$2.16 & 0.6079$\pm$0.03 & 72.46$\pm$3.67 & 0.7250$\pm$0.03 \\
BIOT & 87.95$\pm$3.52 & 0.8778$\pm$0.03 & 74.34$\pm$3.57 & 0.6121$\pm$0.04 & 69.88$\pm$2.15 & 0.7011$\pm$0.03 \\
EEG2Rep & \textbf{94.13$\pm$2.11} & 0.9413$\pm$0.02 &  80.07$\pm$2.63 & 0.6614$\pm$0.02 & \textbf{73.60$\pm$1.47} & 0.7440$\pm$0.02 \\
\underline{SAMBA (Random)} &{69.66$\pm$4.70}  & 0.7943$\pm$0.02                   & 63.38$\pm$2.33 & {0.6377$\pm$0.02}                      &  {57.21$\pm$1.30} & {0.6033$\pm$0.02}              \\ 
\underline{SAMBA (fine-tuned)}  & 93.24$\pm$1.42 & \textbf{0.9793$\pm$0.01}          & \textbf{80.17$\pm$0.68} & \textbf{0.6845$\pm$0.01}     &  {70.89$\pm$0.84} & \textbf{0.7862$\pm$0.01}      \\
\bottomrule
\end{tabular}
}
\end{table}

On TUAB, SAMBA is pretrained using sequences of 10, 30, 60, and 100 seconds, while the compared results use 10-second inputs (Table~\ref{tab:3}). Performance improves consistently with pretraining sequence length. With only 1.0M parameters, SAMBA-100s achieves the best balanced accuracy (82.64\%) and AUROC (0.9054), showing that long-context pretraining improves in-domain representation learning.

\begin{table} 
\centering
\caption{Performance on TUAB.}
\label{tab:3}
\resizebox{0.6\linewidth}{!}{
\begin{tabular}{l|c|c|c}
\toprule
\textbf{Models} & \textbf{Model Size} & \textbf{Balanced ACC} & \textbf{AUROC} \\ \hline
SPaRCNet \cite{jing2023development} & 0.79M &   78.96$\pm$0.18 &   0.8676$\pm$0.0012  \\
ContraWR \cite{yang2023self} & 1.6M & 77.46$\pm$0.41 & 0.8456$\pm$0.0074 \\
CNN-LSTM \cite{li2022motor}  & 2.4M      & 78.48$\pm$0.38          & 0.8569$\pm$0.0051 \\ 
CNN-Transformer \cite{peh2022transformer} & 3.2M              & 77.77$\pm$0.22          & 0.8461$\pm$0.0013 \\ 
BIOT \cite{yang2024biot}  & 3.2M      & 79.59$\pm$0.57          & 0.8815$\pm$0.0043 \\ 
ST-Transformer \cite{song2021transformer} & 3.5M & 79.66$\pm$0.23 & 0.8707$\pm$0.0019 \\
EEGPT \cite{wangeegpt}     & 25M                  & 79.83$\pm$0.30          & 0.8718$\pm$0.0050 \\
LaBraM-Base \cite{jiang2024large} & 5.8M & 81.40$\pm$0.16 & 0.9022$\pm$0.0009 \\

\hline
SAMBA-10s & 1.0 M & 81.50$\pm$0.37 &0.8887$\pm$0.0004\\ 
SAMBA-30s & 1.0 M & 81.88$\pm$0.09 & 0.8937$\pm$0.0025\\ 
SAMBA-60s & 1.0 M &82.18$\pm$0.12& 0.8962$\pm$0.0019\\ 
SAMBA-100s & 1.0 M & \textbf{82.64$\pm$0.17} & \textbf{0.9054$\pm$0.0044} \\ 
\bottomrule
\end{tabular}
}
\end{table}

\subsection{Cross-Domain Transferability}
We evaluate SAMBA's transferability across sequence lengths, datasets, and montages.

\begin{figure}[t]
  \centering
  \subfigure[{\scriptsize Eyes close/open (Accuracy).}]{\label{fig:5a}%
    \includegraphics[width=0.48\linewidth]{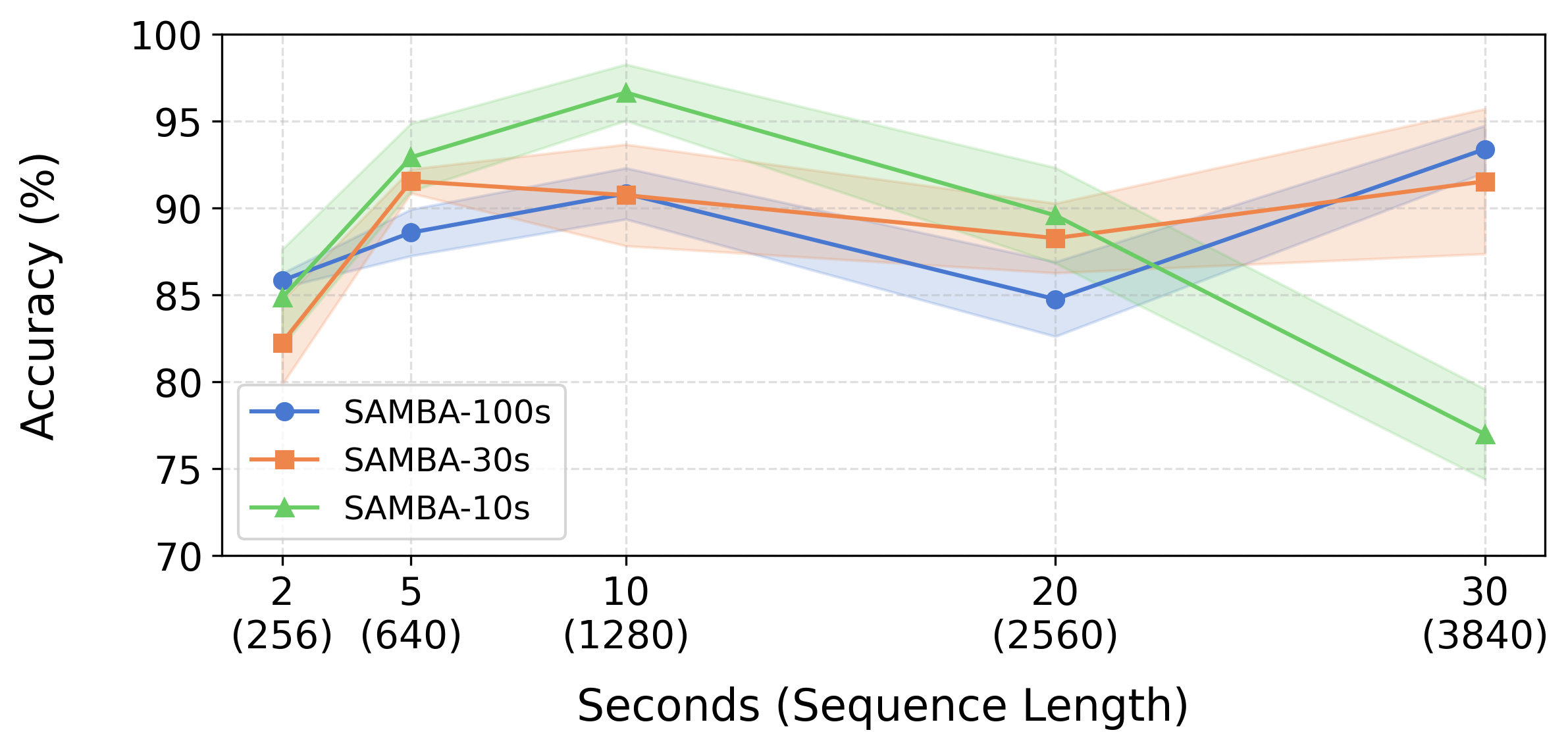}}%
  \hfill
  \subfigure[{\scriptsize Eyes close/open (AUROC).}]{\label{fig:5b}%
    \includegraphics[width=0.48\linewidth]{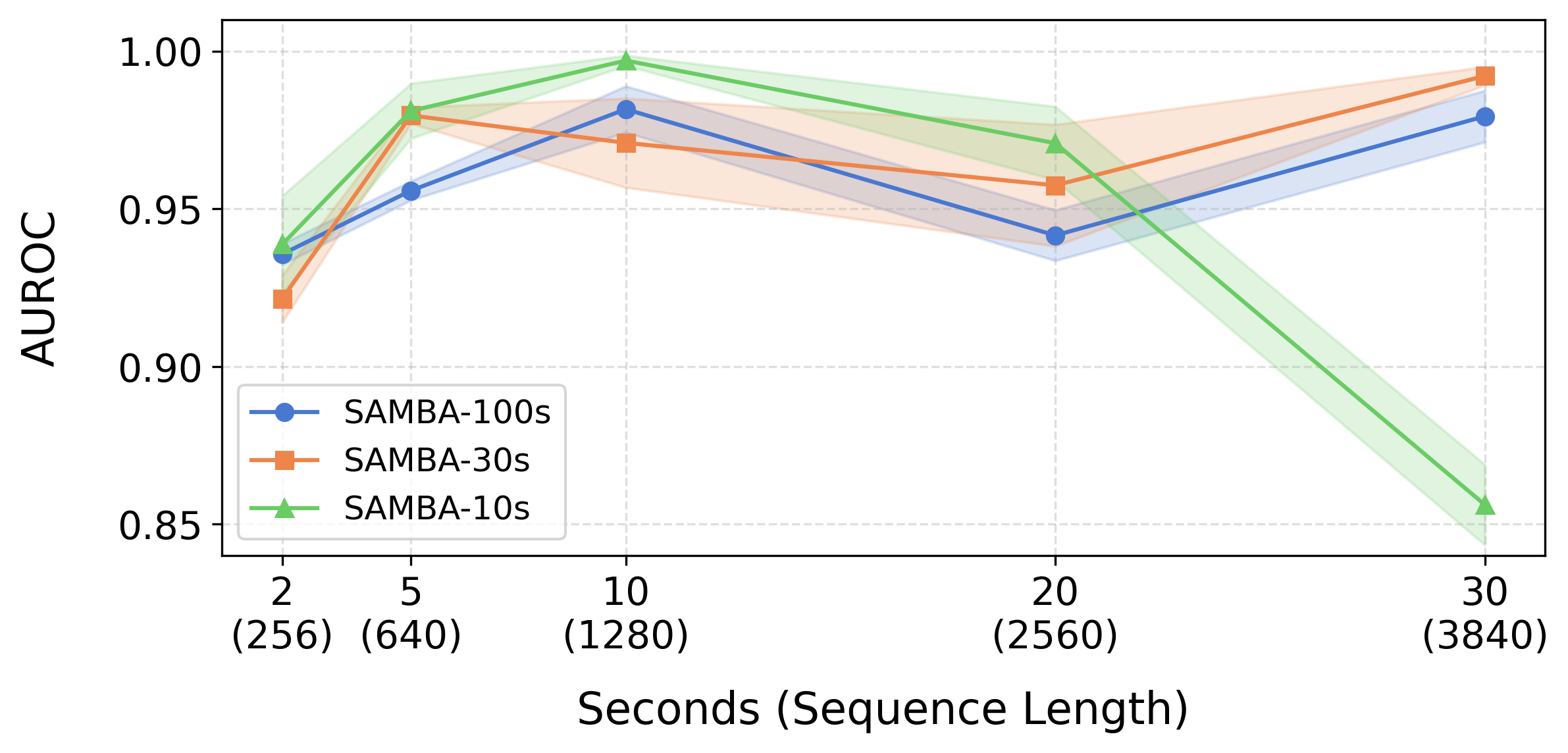}}%
  \vspace{-0.15in}
  \caption{Temporal transferability from long-context pretraining to short-context tasks.}
  \label{fig:5}
\end{figure}

\paragraph{Temporal transferability.}
We evaluate SAMBA models pretrained on 10-, 30-, and 100-second TUAB sequences on the Crowdsourced task using different input durations at 128 Hz. As shown in Figure~\ref{fig:5}, SAMBA-100s maintains stable accuracy and AUROC across input durations and transfers effectively to shorter sequences. Although duration-matched models occasionally achieve the highest AUROC, models pretrained on short sequences generalize poorly to longer inputs. These results show that long-context pretraining provides more transferable temporal representations.

\begin{table}
  \centering
  \caption{Spatial transfer from TUAB to Emotiv datasets with a different montage.}
  \label{tab:4}
  \resizebox{0.68\linewidth}{!}{
  \begin{tabular}{l cc cc cc} 
  \toprule
  \multirow{2}{*}{\textbf{Initialization}} & \multicolumn{2}{c}{\textbf{Crowdsourced}} & \multicolumn{2}{c}{\textbf{DriverDistraction}} & \multicolumn{2}{c}{\textbf{STEW}} \\ \cline{2-7}
    & \textbf{ACC} & \textbf{AUROC} & \textbf{ACC} & \textbf{AUROC} & \textbf{ACC} & \textbf{AUROC}  \\ \hline
  Random & 69.66 & 0.7943 & 63.38 & 0.6377 & 57.21 & 0.6033  \\ 
  In-domain & ($\uparrow$ 23.58) & ($\uparrow$ 0.1850) & ($\uparrow$ 16.80) & ($\uparrow$ 0.0468) & ($\uparrow$ 13.68) & ($\uparrow$ 0.1829) \\
  Cross-domain    & ($\uparrow$ 16.17) & ($\uparrow$ 0.1414) & ($\uparrow$ 10.66) & ($\uparrow$ 0.0313) & ($\uparrow$ 16.39) & ($\uparrow$ 0.1951) \\
  \bottomrule
  \end{tabular}
  }
\end{table}

\paragraph{Spatial transferability.}
We further transfer SAMBA-100s pretrained on TUAB (16 channels and 100-second inputs) to three Emotiv tasks (14 channels and 2-second inputs). Table~\ref{tab:4} reports improvements over random initialization. Cross-domain pretraining improves both accuracy and AUROC on all three tasks despite changes in datasets, tasks, sequence lengths, and electrode montages. These results demonstrate that CDSE supports transfer across heterogeneous EEG configurations.

\subsection{SAMBA as a Foundation Model}
\label{re:foundation}

\begin{table}
\centering
\caption{Linear-probing performance across EEG datasets.}
\label{tab:5}
\resizebox{\linewidth}{!}{
\begin{tabular}{l|cccc|cccccccccccccc}
\hline
\multirow{2}{*}{\textbf{Dataset}} & \textbf{\# of} & \textbf{\# of} & \textbf{Sampling}  & \textbf{Time} & \multicolumn{2}{c}{\textbf{SAMBA-E}} & \multicolumn{2}{c}{\textbf{SAMBA-T}} & \multicolumn{2}{c}{\textbf{EEGPT}} & \multicolumn{2}{c}{\textbf{LaBraM}} & \multicolumn{2}{c}{\textbf{BIOT}}  & \multicolumn{2}{c}{\textbf{MMM}} & \multicolumn{2}{c}{\textbf{CBraMod}} \\
                                  & \textbf{Class} & \textbf{Chan.} & \textbf{Rate (Hz)} & \textbf{(s)}  & Acc.             & W-F1              & Acc.             & W-F1              & Acc.            & W-F1             & Acc.             & W-F1             & Acc.            & W-F1 & Acc.            & W-F1 & Acc.            & W-F1            \\ \hline
\textbf{PhysionetMI}                     & 5              & 64             & 160                & 3             & 28.29            & 0.3070            & 27.41            & 0.3006            & 23.77           & 0.2631           & \textbf{35.79}   & \textbf{0.3575}  & 27.14           & 0.2832  & 19.33 & 0.2229  &   20.32 & 0.2983        \\
\textbf{GrosseWentrup}                     & 2              & 128            & 500                & 7             & 53.50            & 0.5349            & \textbf{57.83}   & \textbf{0.5781}   & 50.67           & 0.5037           & 53.67            & 0.5344           & 50.33           & 0.4530 & 50.83 & 0.5080   & 51.17 & 0.5098        \\
\textbf{BNCI2014-001}                     & 4              & 22             & 250                & 4             & \textbf{32.00}   & \textbf{0.3220}   & 30.57            & 0.3033            & 25.41           & 0.2500           & 28.53            & 0.2807           & 27.31           & 0.2388   &   24.71 & 0.2461  &  25.98 & 0.1436    \\
\textbf{P300-A}                   & 2              & 64             & 240                & 0.8           & 56.26            & 0.6181            & \textbf{58.49}   & \textbf{0.6366}   & 55.79           & 0.6129           & 54.15 & 0.6234    & 49.48 & 0.6117  & 55.20 & 0.6102      &   49.88 & 0.6254\\
\textbf{P300-B}                   & 2              & 64             & 240                & 0.8           & 57.76            & 0.6303            & \textbf{59.22}   & \textbf{0.6424}   & 54.00           & 0.5986           & 56.17 & 0.6326 & 50.88 & \textbf{0.6868}  &53.32 & 0.5853      &   50.49 & 0.6532 \\
\textbf{P300-C}                   & 2              & 64             & 240                & 0.8           & \textbf{61.68}   & \textbf{0.6615}   & 61.28            & 0.6593            & 57.36           & 0.6262           & 58.08 & 0.6565  & 51.40 & 0.6374  &55.74 & 0.6403      &   51.24 & 0.3711 \\ \hline
\end{tabular}
}
\end{table}

Table~\ref{tab:5} compares SAMBA with five EEG foundation models using linear probing on six datasets with different tasks, montages, sampling rates, and durations. All results are obtained from publicly released checkpoints, and the best result for each dataset and metric is shown in bold. SAMBA-E is pretrained on the Emotiv corpus (311k samples, six tasks, and 170 hours), while SAMBA-T is pretrained on over 1,000 hours of 100-second TUAB recordings. SAMBA-E or SAMBA-T achieves the best result on nine of the twelve dataset-metric combinations. 
Some baselines require input resampling, temporal upsampling, or channel mapping; checkpoint selection, compatibility adaptations, and efficiency details are provided in Appendix~\ref{app:foundation-details}.

\begin{figure}
  \centering
  \subfigure[{\scriptsize Memory Usage (MB).}]{%
    \includegraphics[width=0.48\linewidth]{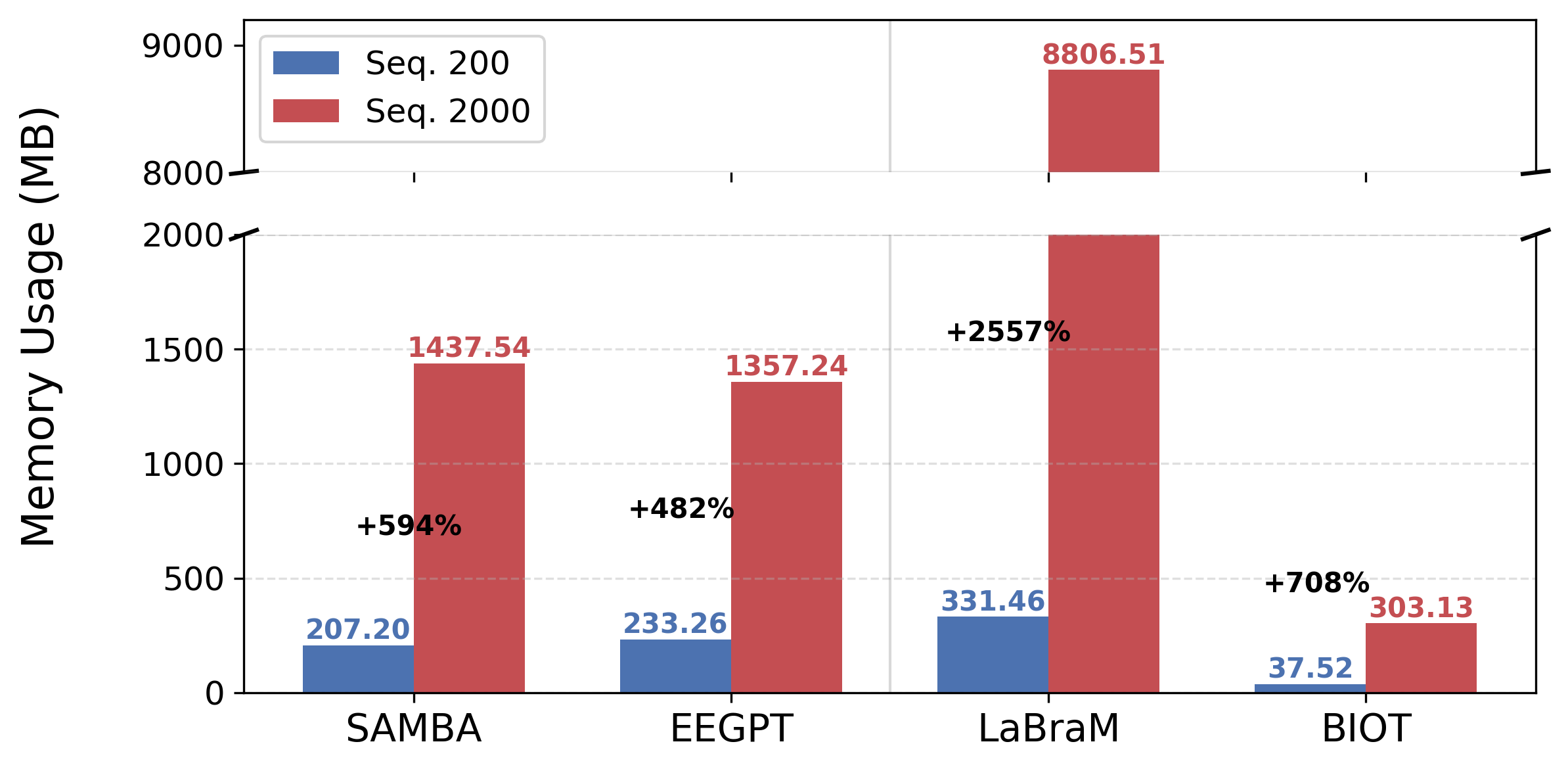}}%
  \hfill
  \subfigure[{\scriptsize Inference Speed (ms).}]{%
    \includegraphics[width=0.48\linewidth]{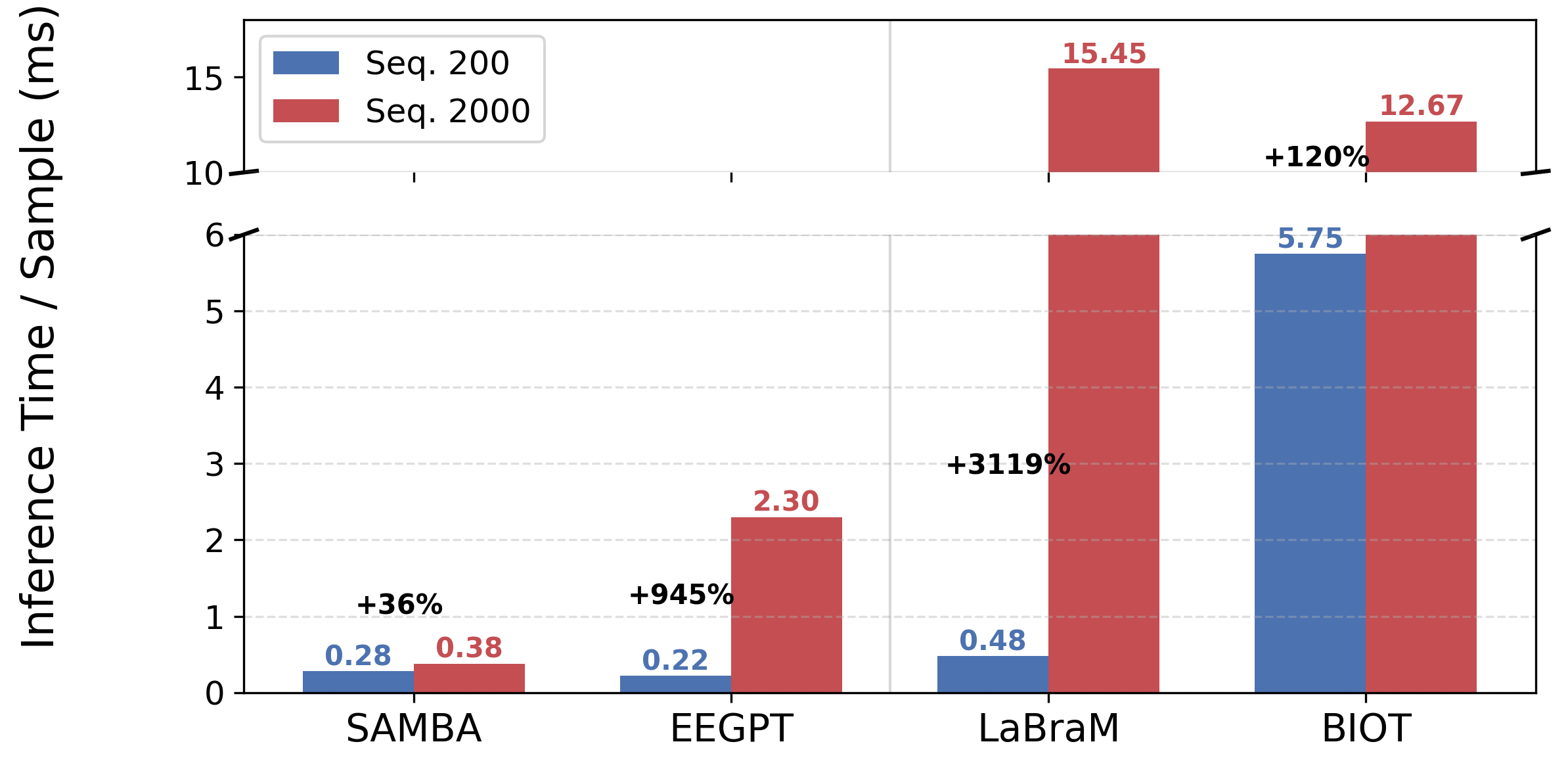}}%
  \vspace{-0.15in}
  \caption{Memory and inference efficiency across sequence lengths.}
  \label{fig:memo}
\end{figure}

\paragraph{Computational efficiency.}
Figure~\ref{fig:memo} compares memory usage and inference time using synthetic 22-channel EEG inputs with 200 and 2{,}000 time steps. Each experiment is repeated five times, and the average is reported; the annotations show the relative increase with sequence length. SAMBA achieves the fastest inference and the second-lowest memory growth. Unlike transformer-based models, SAMBA processes the full input without temporal interpolation while maintaining efficient scaling. 

\section{Conclusion}
We presented \textit{SAMBA}, an efficient EEG representation model for long sequences and heterogeneous electrode layouts. SAMBA supports temporal and spatial transfer without input truncation or channel remapping. Results across thirteen datasets demonstrate its effectiveness and potential as a scalable EEG foundation model.

\clearpage
\newpage
\bibliography{reference}

%%
%% If your work has an appendix, this is the place to put it.
\clearpage
\newpage
\appendix
\small

\section{Representation Visualization}\label{app:moved-tsne}
\begin{figure}[!ht]
  \centering
  \subfigure[{\scriptsize Raw (Acc: 51.30\%)}]{%
    \includegraphics[width=0.245\linewidth]
    {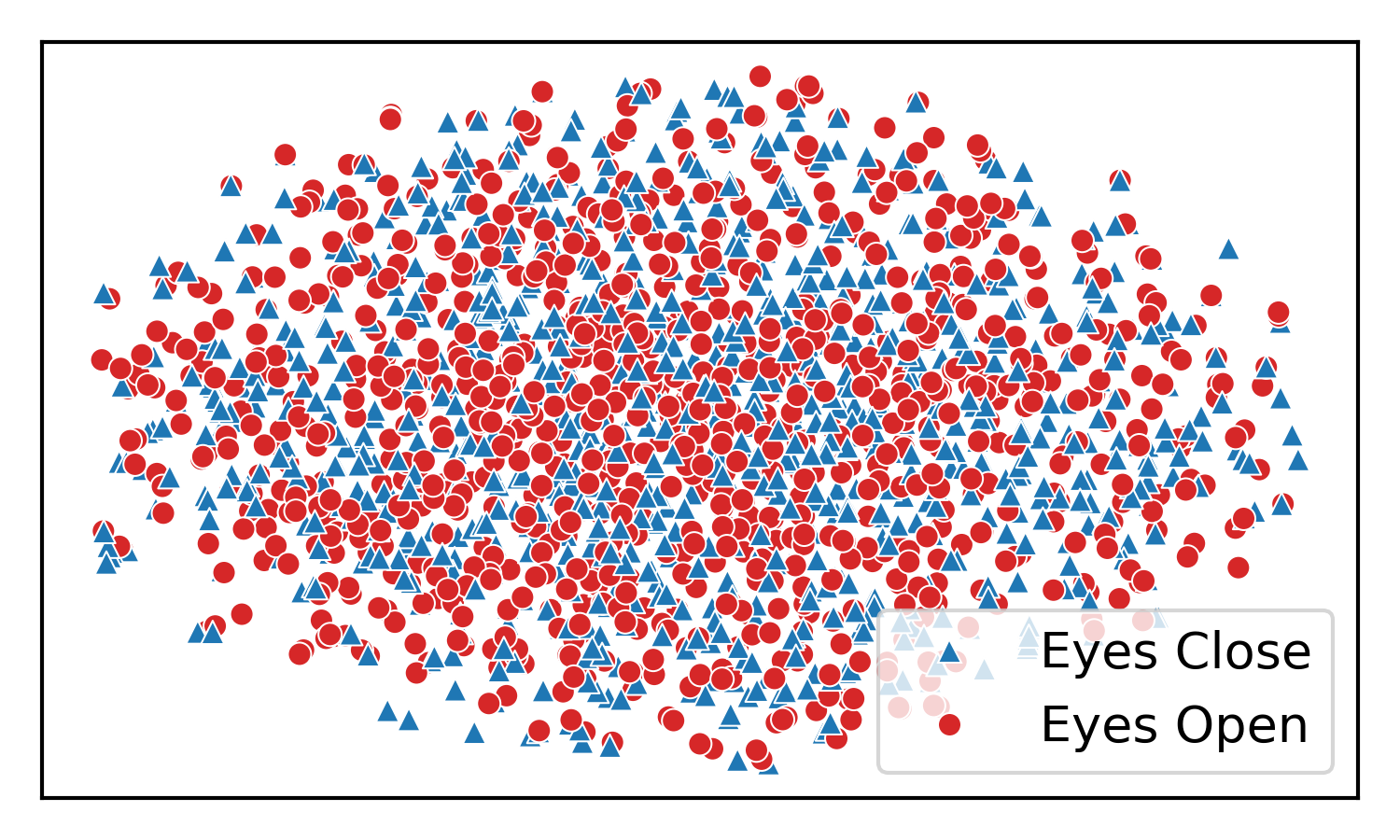}}%
  \hfill
  \subfigure[{\scriptsize SAMBA (Acc: 84.84\%)}]{%
    \includegraphics[width=0.245\linewidth]
    {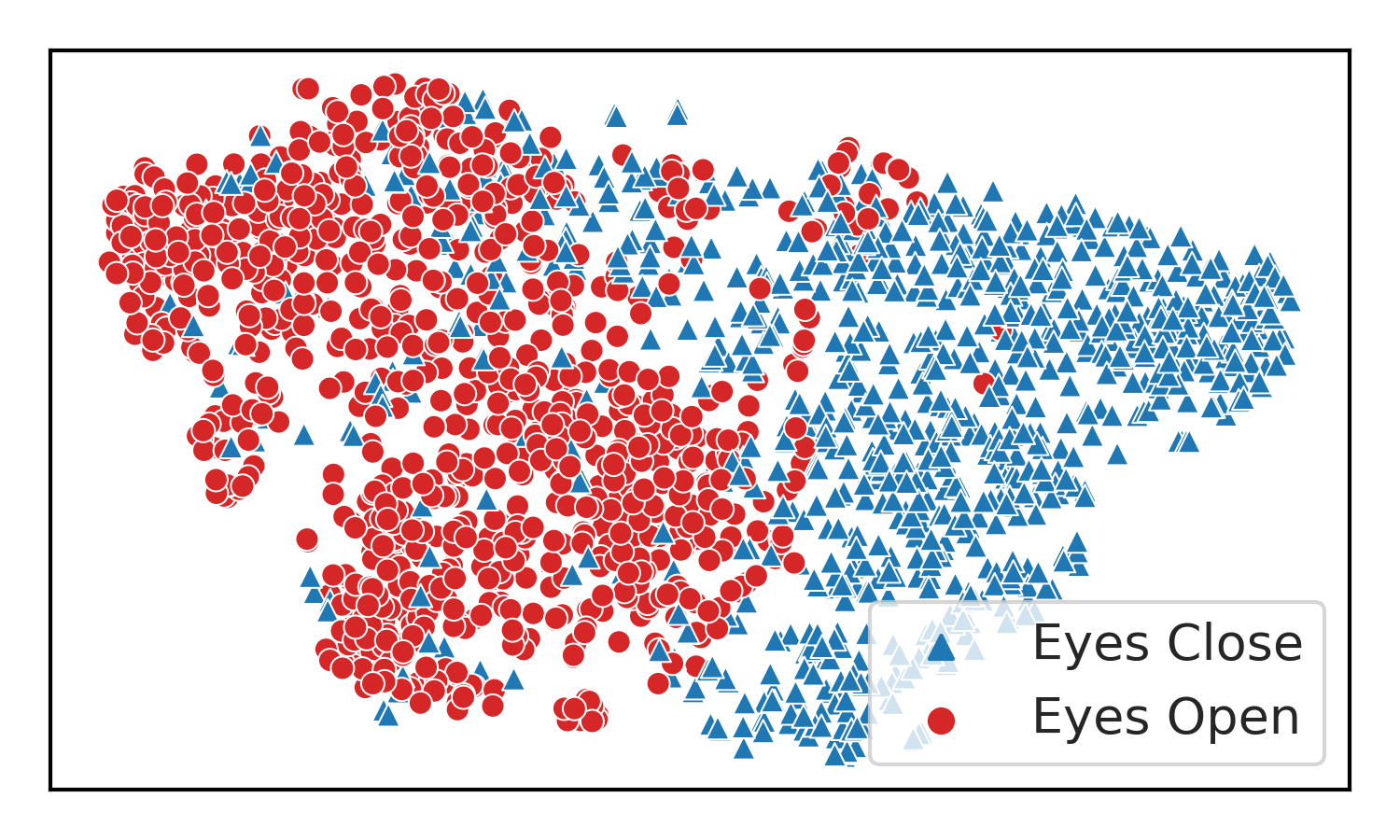}}%
  \hfill
  \subfigure[{\scriptsize Raw (Acc: 59.17\%)}]{%
    \includegraphics[width=0.245\linewidth]
    {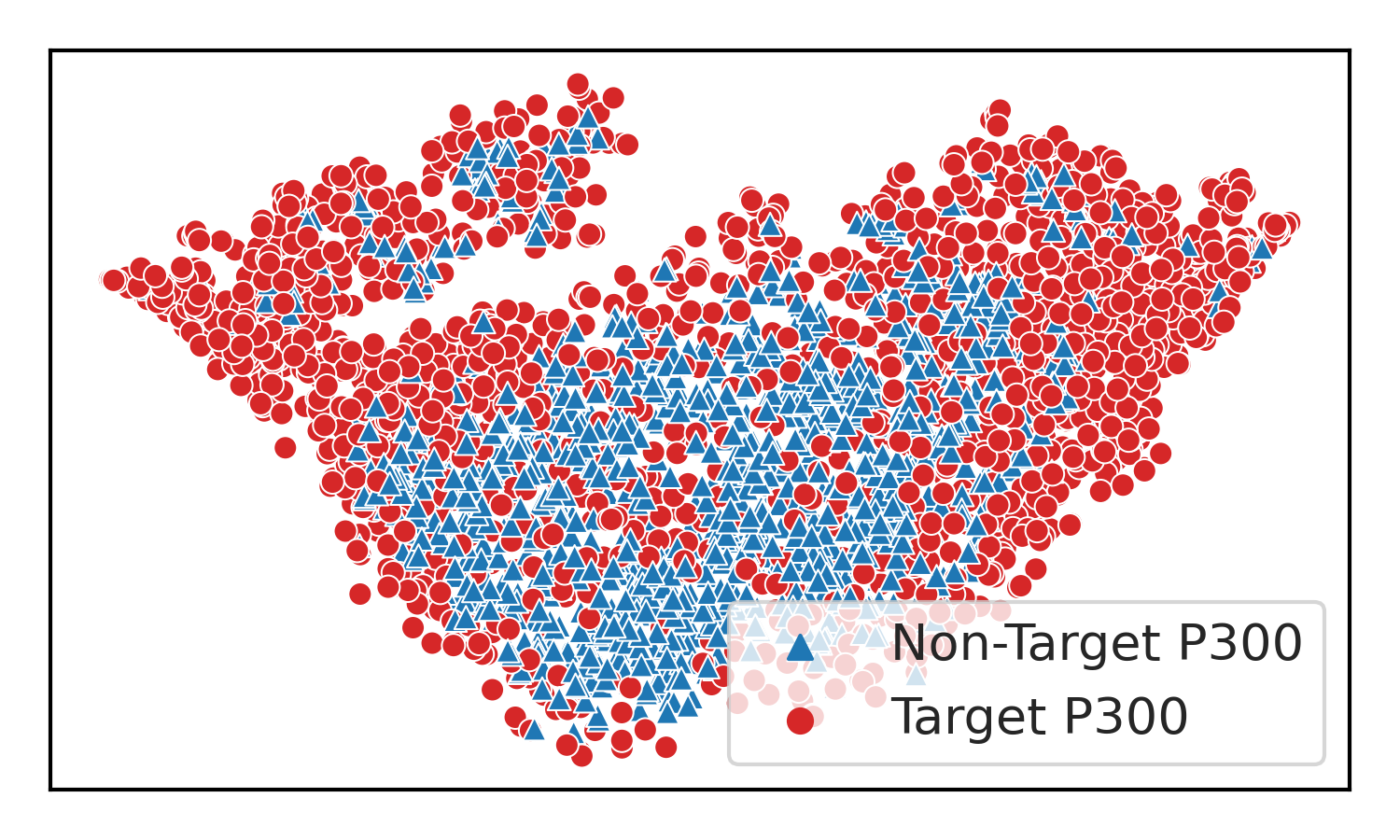}}%
  \hfill
  \subfigure[{\scriptsize SAMBA (Acc: 97.80\%)}]{%
    \includegraphics[width=0.245\linewidth]
    {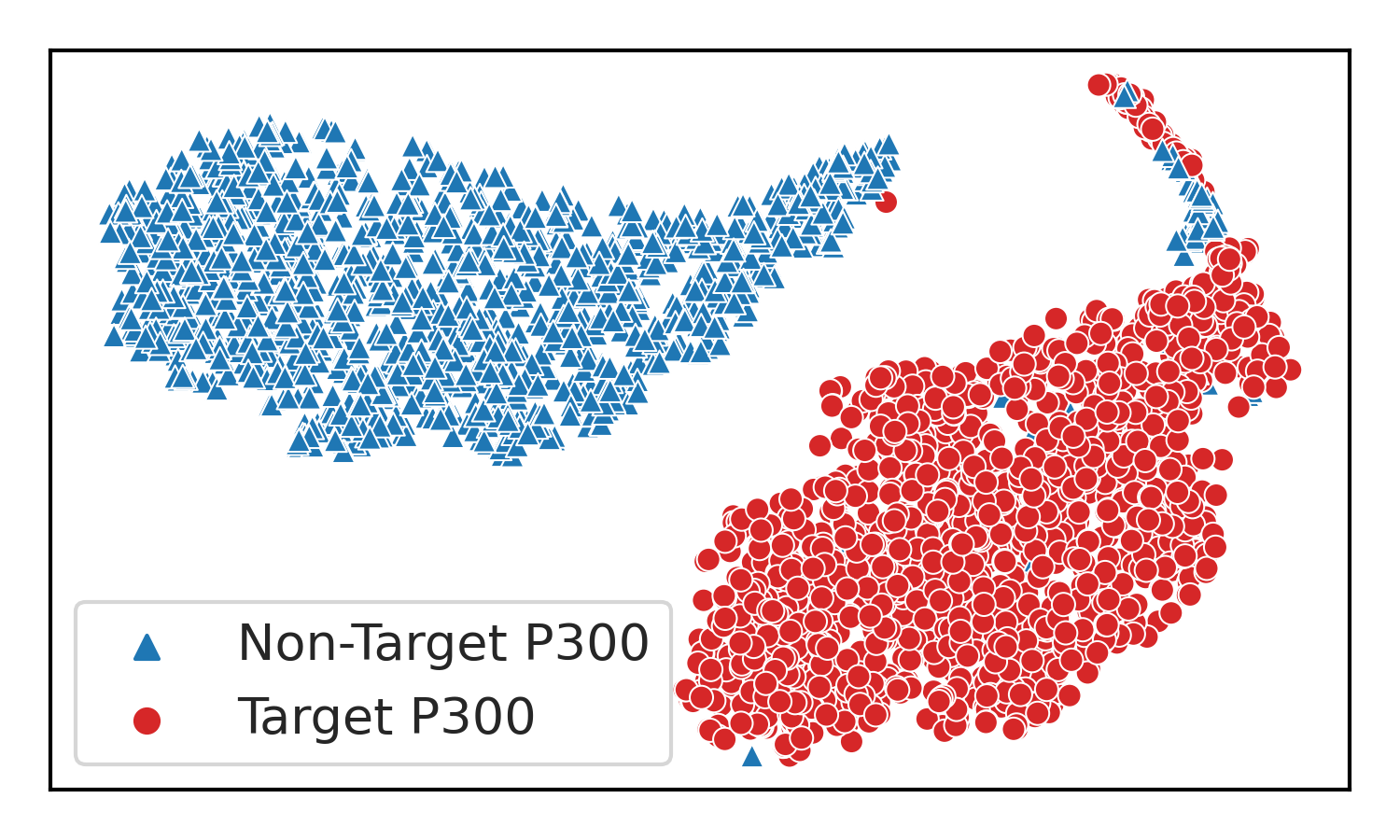}}%
  \vspace{-0.15in}
  \caption{t-SNE plots of raw EEG (a, c) and SAMBA representations (b, d) on Crowdsourced and P300.}
  \label{fig:tSNE}
\end{figure}
Figure~\ref{fig:tSNE} visualizes representations learned by SAMBA across montage and duration shifts. SAMBA is pretrained on TUAB recordings with 16 channels and 100-second sequences (12{,}800 steps at 128~Hz), then transferred to (i) Crowdsourced EEG with 14 channels and 2-second sequences (256 steps at 128~Hz; Appendix~\ref{app:Crowdsourced}) and (ii) P300 EEG with 64 channels and 0.8-second sequences (192 steps at 240~Hz; Appendix~\ref{app:300}). The improved class separation indicates that the learned representations transfer across these spatial and temporal configurations.

\section{Detailed Method Components}\label{app:moved-method}
\subsection{Temporal Semantic Random (TSR) Masking}\label{sec:tsr}

\begin{figure}[!ht]
  \centering
  \includegraphics[width=0.9\linewidth]{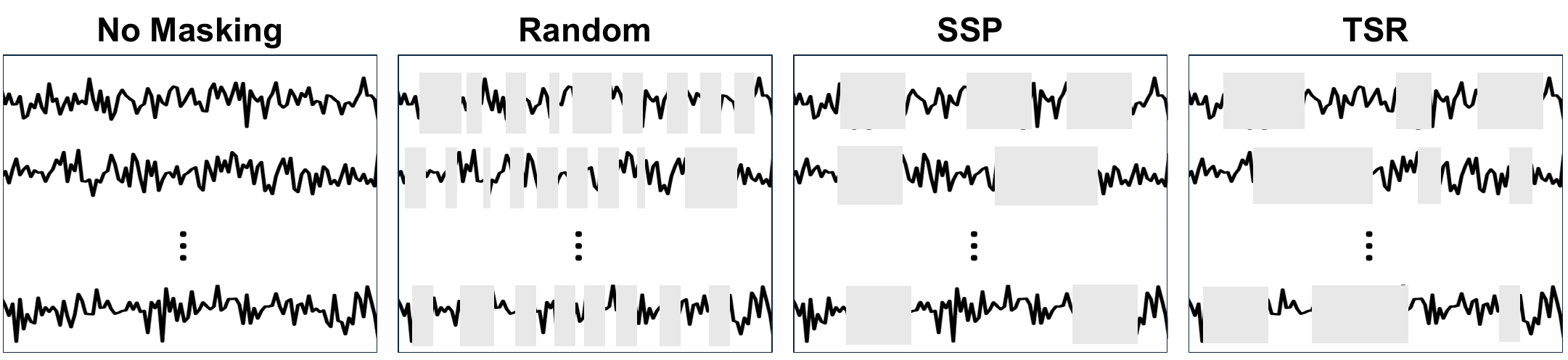}
  \vspace{-0.15in}
  \caption{Comparison of TSR with existing masking strategies.}\label{fig:2}
\end{figure}

Figure~\ref{fig:2} compares masking strategies used in EEG self-supervised learning. Standard random masking~\cite{chien2211maeeg} offers high variability but often disrupts temporal continuity. %Fixed block masking and 
SSP~\cite{eeg2rep2024} attempt to retain semantic coherence by preserving contiguous segments, but overlapping blocks in SSP introduce redundancy and repetitive patterns, limiting temporal diversity, especially in long EEG sequences.

To address these issues, we propose Temporal Semantic Random (TSR) masking. TSR preserves a fixed number of non-overlapping blocks with variable lengths, sampled from a scaled uniform distribution. This design promotes semantic diversity without redundancy while preserving a constant number of visible time steps.

Let \( l \) be the sequence length and \( \rho \) the masking ratio. TSR preserves \( (1 - \rho) \cdot l \) time steps, divided into \( \beta \) non-overlapping blocks. Each block length is sampled as:

\begin{equation}
\forall i \in [1, \beta - 1],\quad
\text{Block}_i \sim \mathcal{U} \left( 
    \left\lfloor \alpha_{\min} \cdot \frac{(1 - \rho) \cdot l}{\beta} \right\rfloor,\ 
    \left\lceil \alpha_{\max} \cdot \frac{(1 - \rho) \cdot l}{\beta} \right\rceil 
\right)
\end{equation}

\begin{equation}
    \text{Block}_{\beta} = (1 - \rho) \cdot l - \sum_{i=1}^{\beta - 1} \text{Block}_i
\end{equation}

All blocks are strictly disjoint to avoid overlap and ensure uniform temporal coverage. Compared to prior methods, TSR better balances semantic continuity and temporal diversity, benefiting modeling long-range EEG dependencies.

\section{State Space Models (SSMs) and Mamba}\label{app:mamba}
Transformer-based models exhibit $\mathcal{O}(n^2)$ complexity in sequence length, making them inefficient for long EEG signals. State Space Models (SSMs), particularly the Mamba series~\cite{gu2023mamba,dao2024transformers}, provide a scalable alternative with linear-time sequence processing. These models evolve a latent hidden state using a recurrence of the form:
\begin{equation}
\mathbf{h}_t = A_t \mathbf{h}_{t-1} + B_t \mathbf{x}_t, \quad
\mathbf{y}_t = C_t \mathbf{h}_t,
\end{equation}
where $\mathbf{x}_t$ is the input at time $t$, $\mathbf{h}_t$ is the internal state, and $A_t$, $B_t$, $C_t$ are learnable matrices. This formulation enables SSMs to process sequences efficiently while maintaining the ability to capture long-range dependencies.

Mamba2~\cite{dao2024transformers} introduces a refined variant of SSMs through the Structured State Space Duality (SSD) framework. This dual formulation expresses SSMs as structured matrix multiplications, specifically, as semiseparable matrices, which admit both a linear recurrent form and a quadratic attention-like form. The structured matrix $M \in \mathbb{R}^{T \times T}$ corresponding to an SSM transformation is given by:
\begin{equation}
M_{tj} = C_t^\top A_{t} \cdots A_{j+1} B_j, \quad \mathbf{y} = M \mathbf{x},
\end{equation}
revealing that SSMs share a mathematical foundation with attention mechanisms, but with improved efficiency.

To reduce hardware inefficiency in early SSM designs, Mamba2 incorporates an SSD algorithm that leverages structured matrix representations and grouped parallel projections. This design enables better utilization of matrix multiplication units on modern accelerators, making it up to $2$–$8\times$ faster than prior implementations. Additionally, Mamba2 introduces analogs to multi-head attention by decomposing the input into multiple parallel SSM channels, enhancing model expressivity without compromising efficiency.

Despite these advances, research in vision and language modeling has shown that SSMs still lag behind comparably sized Transformer models in performance~\cite{lenz2025jamba, zuo2022efficient, glorioso2024zamba}. To address this gap, hybrid architectures have been proposed that combine SSM blocks with Transformer blocks to leverage the strengths of both~\cite{lenz2025jamba}. Other approaches, inspired by the overall architecture of Transformers, suggest stacking SSM blocks and integrating them with attention and MLP layers~\cite{dao2024transformers, glorioso2024zamba}, aiming to enhance the ability to capture local patterns in SSM-based architectures. While these hybrid models strive to achieve competitive performance, they often compromise the core advantage of SSMs: low model complexity.

Unlike convolutions with fixed receptive fields or recurrent models with sequential bottlenecks, Mamba2 supports both global context and dynamic temporal weighting. This makes it particularly suitable for EEG sequences, which often involve complex dependencies across varied time scales. In our framework, Mamba2 is integrated into a convolutional encoder-decoder architecture (see Section~\ref{sec:method}), where each Mamba block processes transposed inputs of shape $(B, T, C)$ and returns to $(B, C, T)$ to maintain spatial-temporal alignment. This integration provides the dual benefits of low-latency computation and enhanced long-term temporal modeling, essential for EEG-based applications.

\section{Time-Frequency Loss Function}\label{app:loss}
SAMBA adopts a reconstruction-based pretraining objective to learn informative latent EEG representations. To preserve both the temporal waveform and spectral structure of EEG signals, we utilized a \textit{Time-Frequency Loss Function}, which combines a Mean Absolute Error (L1) loss in the time domain with a spectral loss computed in the frequency domain.

Given a predicted EEG signal \(\hat{y} \in \mathbb{R}^{B \times T}\) and the corresponding ground truth signal \(y \in \mathbb{R}^{B \times T}\), the total loss is defined as:
\begin{equation}
\mathcal{L}_{\text{TF}} = \alpha \cdot \mathcal{L}_{\text{L1}} + \beta \cdot \mathcal{L}_{\text{Spec}},
\end{equation}
where \(\alpha\) and \(\beta\) are weighting coefficients, and we use \(\alpha = \beta = 1\) in all experiments.

\subsection{L1 Loss (Time-Domain Reconstruction)}
The first term, \(\mathcal{L}_{\text{L1}}\), measures the Mean Absolute Error (MAE) between the predicted and ground truth waveforms:
\begin{equation}
\mathcal{L}_{\text{L1}} = \frac{1}{B T} \sum_{i=1}^{B} \sum_{t=1}^{T} \left| y_{i,t} - \hat{y}_{i,t} \right|.
\end{equation}
Compared to Mean Squared Error (L2 loss), L1 loss penalizes large errors less aggressively, encouraging stable reconstruction and better generalization~\cite{mazilu2011l1}. To compensate for its non-smooth gradients, we employ a OneCycle learning rate schedule during training.

\subsection{Spectral Loss (Frequency-Domain Reconstruction)}
The second term, \(\mathcal{L}_{\text{Spec}}\), encourages accurate spectral reconstruction by minimizing the squared difference between the real-valued Fourier spectra:
\begin{equation}
\mathcal{L}_{\text{Spec}} = \frac{1}{B T} \sum_{i=1}^{B} \sum_{j=1}^{T/2+1} \left| \mathcal{F}(y_i)_j - \mathcal{F}(\hat{y}_i)_j \right|^2,
\end{equation}
where \(\mathcal{F}(\cdot)\) denotes the real-valued discrete Fourier transform (rFFT) along the temporal dimension. Unlike handcrafted frequency band loss, this term compares the full spectrum directly, allowing the model to preserve global oscillatory properties of EEG such as alpha, beta, and theta rhythms.

\section{Experimental Details}\label{app:moved-experimental}
\subsection{Datasets}

\begin{table}
\centering
\caption{Summary of EEG datasets used for evaluation.}
\label{tab:1}
\resizebox{0.9\linewidth}{!}{%
\begin{threeparttable}
\begin{tabular}{@{}lllcrrr@{}}
\toprule
\shortstack{\textbf{Dataset}\\\textbf{Source}} &
\shortstack{\textbf{EEG}\\\textbf{Task}} &
\textbf{Dataset} &
\shortstack{\textbf{\# of}\\\textbf{Channels}} &
\shortstack{\textbf{Sampling}\\\textbf{Rate (Hz)}} &
\shortstack{\textbf{Time}\\\textbf{Steps}} &
\textbf{Samples} \\
\midrule
\multirow{6}{*}{Emotiv\tnote{a}}
& Emotion & DREAMER & 14 & 128 & 256 & 17{,}246 \\
& Eye State & Alpha & 14 & 128 & 256 & 11{,}866 \\
& Eye State & Crowdsourced & 14 & 128 & 256 & 12{,}296 \\
& Mental Workload & STEW & 14 & 128 & 256 & 28{,}512 \\
& Distraction & DriverDistraction & 14 & 128 & 256 & 66{,}197 \\
& Attention & Attention & 14 & 128 & 256 & 21{,}894 \\
\midrule
\multirow{4}{*}{TUAB\tnote{b}}
& \multirow{4}{*}{Abnormal Detection} & TUAB-10s & 16 & 128 & 1{,}280 & 409{,}455 \\
& & TUAB-30s & 16 & 128 & 3{,}840 & 135{,}702 \\
& & TUAB-60s & 16 & 128 & 7{,}680 & 56{,}290 \\
& & TUAB-100s & 16 & 128 & 12{,}800 & 39{,}810 \\
\midrule
\multirow{3}{*}{MOABB\tnote{c}}
& \multirow{3}{*}{Motor Imagery} & PhysionetMI & 64 & 160 & 480 & 18{,}440 \\
& & GrosseWentrup & 128 & 500 & 3{,}500 & 3{,}000 \\
& & BNCI2014-001 & 22 & 250 & 750 & 2{,}592 \\
\midrule
\multirow{3}{*}{BCIC\tnote{d}}
& \multirow{3}{*}{P300 ERP} & P300-A & 64 & 240 & 192 & 33{,}000 \\
& & P300-B & 64 & 240 & 192 & 33{,}000 \\
& & P300-C & 64 & 240 & 192 & 13{,}140 \\
\bottomrule
\end{tabular}%
\begin{tablenotes}[flushleft]
\scriptsize
\setlength{\itemsep}{0pt}
\setlength{\topsep}{1pt}
\item[a] Emotiv datasets follow EEG2Rep~\cite{eeg2rep2024}.
\item[b] TUAB data and preprocessing follow~\cite{lopez2015automated,yang2024biot}.
\item[c] Motor imagery datasets are accessed through MOABB~\cite{moabb2018}.
\item[d] P300 datasets are from the BCI Competitions~\cite{schalk2004bci2000dataset,krusienski2006bciiii}.
\end{tablenotes}
\end{threeparttable}
}
\end{table}

Table~\ref{tab:1} summarizes the thirteen EEG datasets spanning eight cognitive tasks used in this study. Emotiv Research\footnote{\url{https://www.emotiv.com/research}} datasets follow the EEG2Rep preprocessing pipeline~\cite{eeg2rep2024}. For TUAB~\cite{lopez2015automated}, we follow the BIOT splits and preprocessing~\cite{yang2024biot}. For continuous TUAB and Crowdsourced recordings~\cite{williams2023crowdsourced}, we evaluate sequence lengths from 10 to 100 seconds (1{,}280 to 12{,}800 time steps). Motor imagery datasets use the MOABB pipeline\footnote{\url{https://moabb.neurotechx.com}}~\cite{moabb2018}. All datasets except the three single-subject P300 sets~\cite{wang2023st} use subject-wise splits. Additional dataset descriptions are provided in Appendix~\ref{app:data}.

% BCI Competition dataset The BCI Competition datasets, BCIC34a~\cite{blankertz2006bci} and BCIC42a~\cite{brunner2008bci} are used only for downstream evaluation, as their total number of samples is fewer than 3000. Additional details on dataset preprocessing and experimental setup can be found in Appendix~\ref{app:data}.

\subsection{Implementation}
SAMBA is trained in two stages.
% (1) self-supervised pretraining via a temporal masking reconstruction objective without labels, and (2) downstream evaluation through fine-tuning and probing with labels. 
\textit{(1) Pretraining.} SAMBA is pretrained without labels to reconstruct EEG sequences masked by TSR (50\% time and 0\% channels). We train for 200 epochs using AdamW with a weight decay of \(1 \times 10^{-2}\). The batch size is 64 for in-domain evaluation (Section~\ref{re:indomain}) and 4{,}096 for Emotiv foundation-model pretraining (Section~\ref{re:foundation}). A OneCycle schedule~\cite{smith2019super} uses a maximum learning rate of \(5 \times 10^{-4}\), an initial rate of \(2.5 \times 10^{-4}\), a final rate of \(5 \times 10^{-6}\), and 10\% warm-up followed by cosine annealing. \textit{(2) Downstream tasks.} For linear probing, a logistic regression classifier is trained on frozen representations extracted immediately after MDM. For fine-tuning, the pretrained model and an MLP head are jointly optimized for fewer than five epochs. All experiments use 32-bit floating-point precision on two NVIDIA RTX 6000 Ada GPUs.

\subsection{Representation Extraction for Probing}\label{app:representation}
For efficient linear probing on long sequences, we extract hidden representations $\mathbf{F} \in \mathbb{R}^{B \times C \times T'}$ immediately after MDM using a forward hook. We summarize $\mathbf{F}$ along the temporal axis using the minimum, maximum, mean, standard deviation, and five quantiles. The resulting representation is
\begin{equation}
    \mathbf{z} = [\min, \max, \mu, \sigma, Q_{0.05}, Q_{0.25}, Q_{0.5}, Q_{0.75}, Q_{0.95}],
\end{equation}
yielding $\mathbf{z} \in \mathbb{R}^{B \times C \times 9}$. For linear probing, $\mathbf{z}$ is flattened to $\mathbb{R}^{B \times (9C)}$; for nonlinear probing, it is passed directly to the MLP.

\subsection{Evaluation Metrics}\label{app:eval}

\noindent \textbullet\ \textbf{Average channel-wise mean squared error (ACMSE)} measures pretraining reconstruction error averaged across EEG channels.

\noindent \textbullet\ \textbf{Balanced accuracy (B-ACC)} is the mean recall across classes and is used for imbalanced downstream tasks.

\noindent \textbullet\ \textbf{AUROC} is the area under the receiver operating characteristic curve and is used for binary classification.

\noindent \textbullet\ \textbf{Weighted F1 (W-F1)} is the class-frequency-weighted harmonic mean of precision and recall and is used for multiclass evaluation.

\subsection{Foundation-Model Evaluation Details}
\label{app:foundation-details}

All models are evaluated by linear probing using publicly released checkpoints. We use the following checkpoints:
EEGPT Large\footnote{\url{https://github.com/BINE022/EEGPT}, accessed July 1, 2025};
LaBraM Base\footnote{\url{https://github.com/935963004/LaBraM}, accessed July 1, 2025};
the largest BIOT checkpoint pretrained on PRESET and SHHS\footnote{\url{https://github.com/ycq091044/BIOT}, accessed July 1, 2025};
MMM Base\footnote{\url{https://seqml.github.io/MMM/}, accessed October 17, 2025};
and the released CBraMod checkpoint\footnote{\url{https://huggingface.co/weighting666/CBraMod/tree/main}, accessed October 17, 2025}.

LaBraM and BIOT require 200-Hz inputs and fixed temporal windows. Their inputs are therefore resampled to 200 Hz, and the 800-ms P300 signals are temporally upsampled to satisfy their minimum input lengths. EEGPT and BIOT also constrain the number of input channels; we apply the channel-mapping procedure provided by EEGPT when necessary. SAMBA requires none of these adaptations.

For the efficiency experiment, memory usage and inference time are measured on synthetic 22-channel EEG inputs with 200 and 2{,}000 time steps. Each experiment is repeated five times and averaged. BIOT reduces memory through STFT preprocessing, whereas EEGPT controls its token count through temporal interpolation. LaBraM applies full attention to channel-wise temporal patches, resulting in greater memory growth for longer inputs. SAMBA processes the complete sequence using Mamba2 blocks without attention or temporal interpolation.

\section{Ablation Studies}
\label{app:moved-ablation}

\subsection{Architecture and Objective Components}
\label{app:component-ablation}
\begin{table}[!ht]
\centering
\caption{Ablation results on TUAB-100s (12{,}800 time steps per trial).}\label{tab:ablation}
\resizebox{0.8\linewidth}{!}{
\begin{tabular}{llccc}
\toprule
\textbf{Model Variant} & \textbf{Modified Component} & \textbf{ACMSE} & \textbf{B-ACC (\%)} & \textbf{AUROC} \\
\midrule
SAMBA & -- & \textbf{$3.42\times10^{-8}$} & \textbf{82.64} & \textbf{0.9054} \\
\midrule
Setting 1 & TSR $\rightarrow$ random mask & $2.17\times10^{-5}$ & 79.79 ($\downarrow$ 2.84) & 0.8629 ($\downarrow$ 0.04) \\
Setting 2 & TSR $\rightarrow$ SSP & $1.52\times10^{-5}$ & 79.27 ($\downarrow$ 3.34) & 0.8689 ($\downarrow$ 0.04) \\
Setting 3 & MDM $\rightarrow$ Mamba2 block & $1.43\times10^{-6}$ & 80.01 ($\downarrow$ 2.63) & 0.8687 ($\downarrow$ 0.04) \\
Setting 4 & Remove residual in MDM & $9.06\times10^{-8}$ & 81.41 ($\downarrow$ 1.23) & 0.8867 ($\downarrow$ 0.02) \\
Setting 5 & Remove $\mathcal{L}_{\text{T}}$ & $2.20\times10^{-7}$ & 80.18 ($\downarrow$ 2.46) & 0.8677 ($\downarrow$ 0.04) \\
Setting 6 & Remove $\mathcal{L}_{\text{F}}$ & $1.47\times10^{-9}$ & 80.54 ($\downarrow$ 2.10) & 0.8827 ($\downarrow$ 0.02) \\
Setting 7 & All Mamba $\rightarrow$ Conv & $5.04\times10^{-7}$ & 78.67 ($\downarrow$ 3.97) & 0.8552 ($\downarrow$ 0.05) \\
Setting 8   & Mamba $\rightarrow$ Attention & OOM & OOM & OOM \\
\bottomrule
\end{tabular}
}
\end{table}

We conduct ablation studies on TUAB-100s using eight configurations:
(1) replacing TSR with random masking,
(2) replacing TSR with SSP masking~\cite{eeg2rep2024},
(3) replacing MDM with a standard Mamba2 block,
(4) removing the residual connection in MDM,
(5) removing the time-domain loss,
(6) removing the frequency-domain loss,
(7) replacing all Mamba modules with convolutional layers, and
(8) replacing Mamba with attention.
We report average channel-wise MSE (ACMSE) during pretraining and balanced accuracy (B-ACC) and AUROC during fine-tuning.

As shown in Table~\ref{tab:ablation}, each architectural modification reduces downstream performance. Replacing TSR with random or SSP masking causes clear decreases in B-ACC and AUROC, demonstrating the benefit of temporal semantic masking. Replacing MDM with a standard Mamba2 block also reduces performance, supporting the use of differential processing. Removing either training objective lowers downstream performance, although removing the frequency-domain loss produces a lower ACMSE. This result indicates that reconstruction error alone does not fully reflect downstream representation quality. The convolutional variant performs worst among the executable variants, while the attention-based variant runs out of memory. These results support the contribution of each component and the efficiency of Mamba for long EEG sequences.

\subsection{Pretraining Diversity and Feature Aggregation}
\label{app:addexperiment}

\begin{table}[!ht]
\centering
\caption{Effects of pretraining diversity and feature aggregation.}\label{tab:7}
\resizebox{0.9\linewidth}{!}{
\begin{tabular}{lcccccccc}
\hline
\multirow{2}{*}{\textbf{Dataset}} & \multicolumn{2}{c}{\textbf{SAMBA-E}} & \multicolumn{2}{c}{\textbf{SAMBA-E (mean)}} & \multicolumn{2}{c}{\textbf{SAMBA*}} & \multicolumn{2}{c}{\textbf{SAMBA* (mean)}} \\
                                  & \textbf{Acc.}   & \textbf{W-F1}    & \textbf{Acc.}          & \textbf{W-F1}           & \textbf{Acc.}    & \textbf{W-F1}    & \textbf{Acc.}           & \textbf{W-F1}           \\ \hline
\textbf{PhysionetMI}              & 28.29           & 0.3070           & \textbf{29.83}         & \textbf{0.3226}         & 27.41            & 0.2985           & 29.04                   & 0.3140                  \\
\textbf{GrosseWentrup}            & 53.50           & 0.5349           & 57.72                  & 0.5772                  & 57.50            & 0.5749           & \textbf{57.78}          & \textbf{0.5776}         \\
\textbf{BNCI2014-001}             & 32.00           & 0.3220           & \textbf{33.28}         & \textbf{0.3285}         & 29.54            & 0.2954           & 32.25                   & 0.3198                  \\
\textbf{P300-A}                   & 56.26           & 0.6181           & 53.35                  & 0.5927                  & \textbf{58.01}   & \textbf{0.6323}  & 56.22                   & 0.6179                  \\
\textbf{P300-B}                   & \textbf{57.76}  & \textbf{0.6303}  & 53.37                  & 0.5929                  & 57.23            & 0.6260           & 54.75                   & 0.6051                  \\
\textbf{P300-C}                   & \textbf{61.68}  & \textbf{0.6615}  & 58.58                  & 0.6377                  & 60.85            & 0.6546           & 57.69                   & 0.6327                  \\ \hline
\end{tabular}
}
\end{table}
Table~\ref{tab:7} examines the effects of pretraining diversity and feature aggregation during linear probing. SAMBA-E is pretrained on six datasets spanning five EEG tasks, with 311{,}011 samples in total. SAMBA* is pretrained on five datasets spanning four tasks, with 140{,}765 samples. The variants without quantile aggregation use mean aggregation instead.

Overall, SAMBA-E outperforms SAMBA* on most downstream datasets, suggesting that a larger and more diverse pretraining corpus improves transferability. Quantile aggregation generally performs better on the P300 datasets, whereas mean aggregation performs better on the motor imagery datasets. These results indicate that the preferred feature aggregation strategy may depend on the downstream EEG paradigm.

\section{Details of Dataset}\label{app:data}
\subsection{BNCI2014-001}

% \begin{figure} [h]
%   \centering
%   \includegraphics[width=0.8\linewidth]{Figure/Fig_bci42a.png}
%   \caption{Timing scheme of a single trial of BCI Competition IV-2a}\label{fig:bci42a}
% \end{figure}
The BNCI2014-001 dataset, also referred to as BCI Competition IV-2a~\cite{brunner2008bci}, contains EEG recordings from nine subjects using 22 Ag/AgCl electrodes placed according to the international 10--20 system
% (the electrode positions can be found in Table~\ref{tab:bci42a} with the numbers in topographic maps), 
with a sampling rate of 250 Hz. The EEG data for each subject were recorded in two sessions on different days, one for the training set and one for the testing set. Each session consisted of six runs with short breaks between them, and each run contained 48 trials. In other words, each session comprised 288 trials across four classes, with each class containing 72 trials. Each class represented a motor imagery task: imagining the movement of the left hand (L), right hand (R), both feet (F), and tongue (T). This resulted in a total of 144 trials for each MI task, with 72 from the training set and 72 from the testing set. The timeline for a trial is approximately 7.5 seconds. 
% as detailed in Figure~\ref{fig:bci42a}. 
At the start of a trial ($t = 0$ s), a fixation cross appears on the black screen. A cue is then displayed at $t = 2$ s for 1.25 seconds. Upon seeing the cue, subjects perform the corresponding MI task until the fixation cross disappears from the screen at $t = 6$ s. A short break follows each trial. Recordings from $t = 3$ s to $t = 6$ s are used for further analysis. 

% \begin{table}
% \centering
% \caption{Dataset II: Correspondence table between electrodes and numbers}
% \label{tab:bci42a}
% \resizebox{\linewidth}{!}{%
% \begin{tabular}{rl rl rl rl rl}
% \toprule
% \textbf{ID} & \textbf{Electrode} & \textbf{ID} & \textbf{Electrode} & \textbf{ID} & \textbf{Electrode} & \textbf{ID} & \textbf{Electrode} & \textbf{ID} & \textbf{Electrode} \\
% \midrule
% 1  & Fz   & 2  & FC3  & 3  & FC1  & 4  & FCz  & 5  & FC2  \\
% 6  & FC4  & 7  & C5   & 8  & C3   & 9  & C1   & 10 & Cz   \\
% 11 & C2   & 12 & C4   & 13 & C6   & 14 & CP3  & 15 & CP1  \\
% 16 & CPz  & 17 & CP2  & 18 & CP4  & 19 & P1   & 20 & Pz   \\
% 21 & P2   & 22 & POz  &     &      &     &      &     &     \\
% \bottomrule
% \end{tabular}
% }
% \end{table}

\subsection{GrosseWentrup2009}

The GrosseWentrup2009 dataset contains motor imagery EEG from 10 healthy subjects recorded with 128 electrodes placed according to the extended 10--20 system~\cite{grosse2009beamforming}. Each subject performed 150 trials of motor imagery for each hand, totaling 300 trials. During each 7-second trial, subjects imagined hand movement following an arrow cue presented after a 3-second fixation. EEG was recorded at 500 Hz with Cz as the reference and rereferenced to the common average offline. No artifact correction or trial rejection was applied. Three-dimensional electrode coordinates were recorded using an ultrasound tracking system. The dataset is available through MOABB~\cite{moabb2018}.

\subsection{PhysioNet MI}
The PhysioNet Motor Imagery dataset~\cite{schalk2004bci2000} includes EEG recordings from 109 subjects performing executed and imagined movements of the hands and feet. EEG was recorded from 64 channels at 160 Hz using the BCI2000 system. Each subject completed 14 runs, and each trial lasted 3 seconds.

\subsection{P300-A, B, C} \label{app:300}

The P300-A and P300-B datasets are from BCI Competition III dataset II~\cite{krusienski2004bci}, and P300-C is from BCI Competition II dataset IIb~\cite{kaper2004bci}. Each dataset contains data from one subject recorded with 64 electrodes at 240 Hz. In the Farwell-Donchin paradigm, participants viewed a $6\times6$ symbol matrix whose rows and columns were randomly intensified at 5.7 Hz. Intensification of the target row or column elicited a P300 response.
% (an example visualized in Fig.~\ref{fig:p300}). 
Each sequence contained 12 intensifications: two targets and ten non-targets.
% (total trial number shown in Table~\ref{tab:p300}). 
Each intensification lasted 100 ms and was followed by a 75-ms blank interval. The sequence was repeated 15 times per target symbol. The P300-C test set contains 31 symbols, while P300-A and P300-B each contain 100 test symbols. For the representation analysis in Figure~\ref{fig:tSNE}, target and non-target trials are separately averaged across repetitions. This changes the original 1:5 class ratio to approximately 1:1 for clearer visualization.

\subsection{Attention Dataset} 
The Attention dataset contains two visual and two auditory tasks designed to measure attention to repeated stimuli. In the visual tasks, participants identified consecutively repeated four-digit numbers over 640 seconds. In the auditory tasks, they identified repeated words over 540 seconds. EEG was recorded using a 14-channel Emotiv EPOC headset. Data were collected from 31 participants, of whom four were excluded because of poor signal quality.

\subsection{Alpha \& Crowdsourced} \label{app:Crowdsourced}
The Alpha and Crowdsourced datasets contain EEG recordings from eyes-open and eyes-closed resting-state tasks. The Alpha data are extracted from the Attention and STEW datasets during resting-state segments where subjects alternated between eyes-open and eyes-closed conditions. The Crowdsourced dataset~\cite{williams2023crowdsourced} includes recordings from 60 participants, among whom 13 completed both conditions using EPOC+, EPOC X, or EPOC devices with 14 channels. EEG signals were originally sampled at 2048 Hz and later downsampled to 128 Hz. Raw EEG data, along with preprocessing and analysis scripts, are publicly available on the Open Science Framework (OSF).

\subsection{DriverDistraction}
DriverDistraction contains EEG from 17 participants during approximately 40 minutes of simulated driving. Distraction tasks included passenger conversation, mobile-phone use, and problem solving. EEG was recorded from 14 channels at 128 Hz using an Emotiv EPOC headset, producing approximately 5.5 million labeled time points.

\subsection{STEW}
The STEW dataset is a public mental-workload dataset containing EEG from 48 participants who completed the SIMKAP multitasking test~\cite{lim2018stew}. Resting-state baseline EEG was also recorded. Each participant contributed 2.5 minutes of 14-channel EEG sampled at 128 Hz and rated perceived workload on a scale from 1 to 9. Ratings above 4 are labeled as high workload, and ratings of 4 or below as low workload.

\subsection{DREAMER}
The DREAMER dataset contains EEG and electrocardiogram recordings from 23 participants during affective stimulation with audiovisual clips~\cite{katsigiannis2017dreamer}. EEG was recorded using a 14-channel Emotiv EPOC device. Participants rated valence, arousal, and dominance after each stimulus. We use only the EEG recordings and arousal labels.

\end{document}